\documentclass[letterpaper]{article} 
\usepackage{aaai2027}  
\usepackage[hyphens]{url}  
\usepackage{graphicx} 
\usepackage{natbib}  
\usepackage{caption} 
\usepackage{algorithm}
\usepackage{algorithmic}

\usepackage{booktabs}
\usepackage[table]{xcolor}

\usepackage{amsmath}
\usepackage{amsfonts}
\usepackage{amssymb}

\newcommand{\papertitle}{DIVE: Dynamic Iterative Visual Evidence Construction for \\
Efficient Vision-Language Models}
\title{\papertitle}
\author{
    Chen Zhong\textsuperscript{\rm 1},
    Xiao An\textsuperscript{\rm 1},
    Zijie Wang\textsuperscript{\rm 1},
    Jiepan Li\textsuperscript{\rm 1},
    Guangyi Yang\textsuperscript{\rm 2},
    Wei He\textsuperscript{\rm 1}\corresponding
}
\affiliations{
    \textsuperscript{\rm 1}State Key Laboratory of Information Engineering in Surveying Mapping and Remote Sensing, Wuhan University, China\\
    \textsuperscript{\rm 2}Electronic Information School, Wuhan University, China
}

\begin{document}

\maketitle

\begin{abstract}

Visual inputs in vision-language models (VLMs) are often encoded into substantially longer token sequences than text, making visual tokens a major bottleneck for efficient inference. Abundant recent methods address this bottleneck by scoring token importance and pruning low-scoring tokens in a single pass. However, one-shot scoring is insufficient because a token's prompt-relevant usefulness depends on the evidence already retained. Motivated by this insight, we introduce DIVE (Dynamic Iterative Visual Evidence Construction), a training-free framework that recasts visual-token pruning as dynamic evidence construction. DIVE repeatedly selects the remaining token with the highest residual-conditioned score, updates the visual and prompt residuals to discount the evidence already explained, and re-evaluates the remaining tokens. This \textit{select--update--re-evaluate} process builds a retained set of complementary, prompt-relevant evidence. Experiments across eight image-understanding benchmarks show that DIVE consistently preserves performance across token budgets. With an 88.9\% reduction in visual tokens, DIVE retains 98.2\% of the uncompressed model's average performance. Code is available at \url{https://github.com/Zhong-Chenchen/DIVE.git}.

\end{abstract}

\section{Introduction}
\label{sec:introduction}

By encoding visual inputs as sequences of visual tokens, Vision-Language Models (VLMs) enable Large Language Models (LLMs) to reason over visual content~\cite{qwen2-vl}. However, dense visual tokenization creates a pronounced imbalance between the visual and textual components of the input. Under a common LLaVA-style encoding setting, a single image produces 576 visual tokens, whereas a prompt such as “What is the person doing?” contains fewer than 10 text tokens~\cite{llava}. Processing these long visual sequences substantially increases the computation and memory required for inference~\cite{FastV,visionzip}. The central challenge for efficient VLM inference is therefore to shorten the visual sequence without discarding the evidence needed to answer the prompt~\cite{rethinking}.

\begin{figure}[t]
    \centering
    \includegraphics[width=\columnwidth]{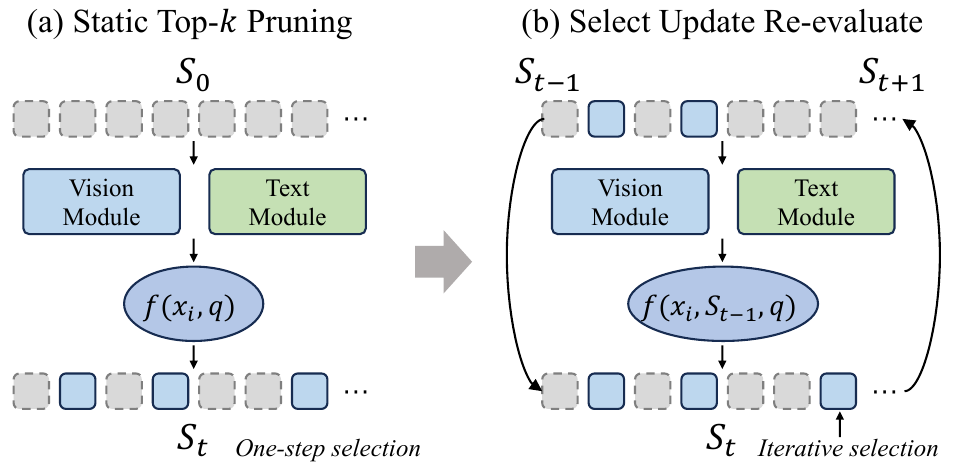}
    \caption{Comparison of two strategies: (a) conventional Static Top-$k$ Pruning and (b) our Select–Update–Re-evaluate Pruning. $S_i$ denotes the set retained after the $i$-th selection.}
    \label{fig:two_paradigms}
\end{figure}

To address this challenge, most visual token pruning methods adopt a \textit{Static Top-$k$ Pruning} paradigm to remove redundant tokens, improving inference efficiency while preserving model performance~\cite{visionzip}. In these methods, tokens are scored using signals such as attention weights~\cite{FastV,Hired}, language relevance~\cite{SparseVLM}, or inter-token redundancy~\cite{dart}, and the tokens are retained under a target budget. Once computed, the token scores and the resulting ranking remain fixed, and the retained subset is determined directly from this ranking. 

As VLMs scale toward longer visual sequences, a structural limitation of \textit{static top-$k$ pruning} becomes increasingly pronounced: each token is assigned a fixed importance score before subset construction, even though its usefulness can depend on which tokens have already been retained~\cite{HoloV,ToMe}. For example, once the selected tokens have sufficiently characterized a salient object, retaining additional high-scoring patches from the same object region contributes little new information~\cite{ViTCoP}. The remaining budget would instead be better allocated to previously uncovered image regions, secondary objects, or complementary contextual details.

Our key insight is that a visual token's value depends not only on its standalone relevance, but also on its prompt-relevant residual state with respect to the retained set. Guided by this insight, we propose DIVE (Dynamic Iterative Visual Evidence Construction), a training-free framework for \textit{dynamic visual token construction}. At each step, DIVE selects the highest-scoring candidate under the current prompt-conditioned state, discounts the visual and textual evidence it explains, and re-scores the remaining candidates. This \textit{select--update--re-evaluate} process allows token scores to evolve with the retained set and guides later selections toward complementary, prompt-relevant evidence. DIVE requires no architectural modification.

Experiments on LLaVA-1.5-7B across eight image-understanding benchmarks show that DIVE achieves the best average performance at every evaluated token budget. Results on LLaVA-1.5-13B, LLaVA-NeXT-7B/13B, and Qwen2-VL further establish its applicability across model scales and VLM architectures. Evaluations on LLaVA-OV-7B extend these findings to video understanding, demonstrating that DIVE remains effective on video benchmarks. Performance--latency comparisons further show that DIVE delivers the highest task performance under comparable end-to-end runtime budgets. Controlled ablations attribute these gains to DIVE's iterative \textit{select--update--re-evaluate} process, which consistently outperforms its frozen-score counterpart. In summary, our main contributions are:

\begin{itemize}
    \item We propose DIVE, a training-free framework that constructs a compact set of visual evidence through set-dependent token selection, without retraining or modifying the underlying VLM.

    \item We design prompt-conditioned residual scoring and coupled one-sided updates that discount covered visual-text evidence, enabling a \textit{select--update--re-evaluate} process that favors complementary, prompt-relevant tokens.

    \item Extensive experiments demonstrate that DIVE achieves strong performance across diverse VLMs, while comprehensive analyses validate its effectiveness in balancing accuracy and efficiency.
\end{itemize}

\section{Related Work}
\label{sec:related_work}

\subsection{Vision-Language Models}

VLMs typically integrate a vision encoder, a modality projector, and an LLM to support multimodal understanding and reasoning~\cite{blip2,instructblip}. Visual encoding strategies determine how token counts scale with image resolution and video length. LLaVA-1.5 adopts fixed-resolution encoding, producing a fixed-length visual-token sequence for each image~\cite{llava}. To preserve finer details, crop-based high-resolution models encode multiple local views, causing the sequence length to grow with the number of crops~\cite{internvl3,llava-next,llava-ov}. Native-resolution models scale token counts with the input dimensions, producing longer sequences for larger images~\cite{qwen2-vl,seed1.5-vl}, while video VLMs further extend visual sequences across multiple frames~\cite{llava-video,videollama3,videoxl-pro}. These increasingly long visual sequences raise prefill latency, attention computation, and KV-cache memory, making efficient visual-token processing a central challenge for high-resolution and long-video understanding.

\subsection{Token Pruning for VLMs}

Visual token pruning reduces VLM inference cost by retaining a subset of visual tokens before or within the LLM~\cite{efficientvlm-survey,tokencompression-survey}. Most existing methods follow a \textit{Static Top-$k$ Pruning} paradigm. FastV and HiRED estimate token utility from visual attention responses~\cite{FastV,Hired}, while SparseVLM ranks tokens according to their relevance to the textual instruction~\cite{SparseVLM}. VisionZip and DART reduce visual redundancy by preserving dominant or non-duplicated tokens~\cite{visionzip,dart}. Recently, some studies have employed greedy algorithms to optimize subset construction through objectives such as maximum coverage~\cite{mmtok} and subspace reconstruction~\cite{resprune}.  Since each decision depends on the retained subset, this process can be viewed as an improvement over \textit{static top-$k$ pruning}. However, these methods do not explicitly model how visual evidence and prompt requirements progressively evolve during selection. To address this gap, we use each selected token as feedback to jointly update the visual and textual residual states, enabling the remaining candidates to be re-evaluated for complementary and prompt-relevant evidence.
\section{Method}
\label{sec:method}

\begin{figure*}[!t]
    \centering
    \includegraphics[width=0.95\textwidth]{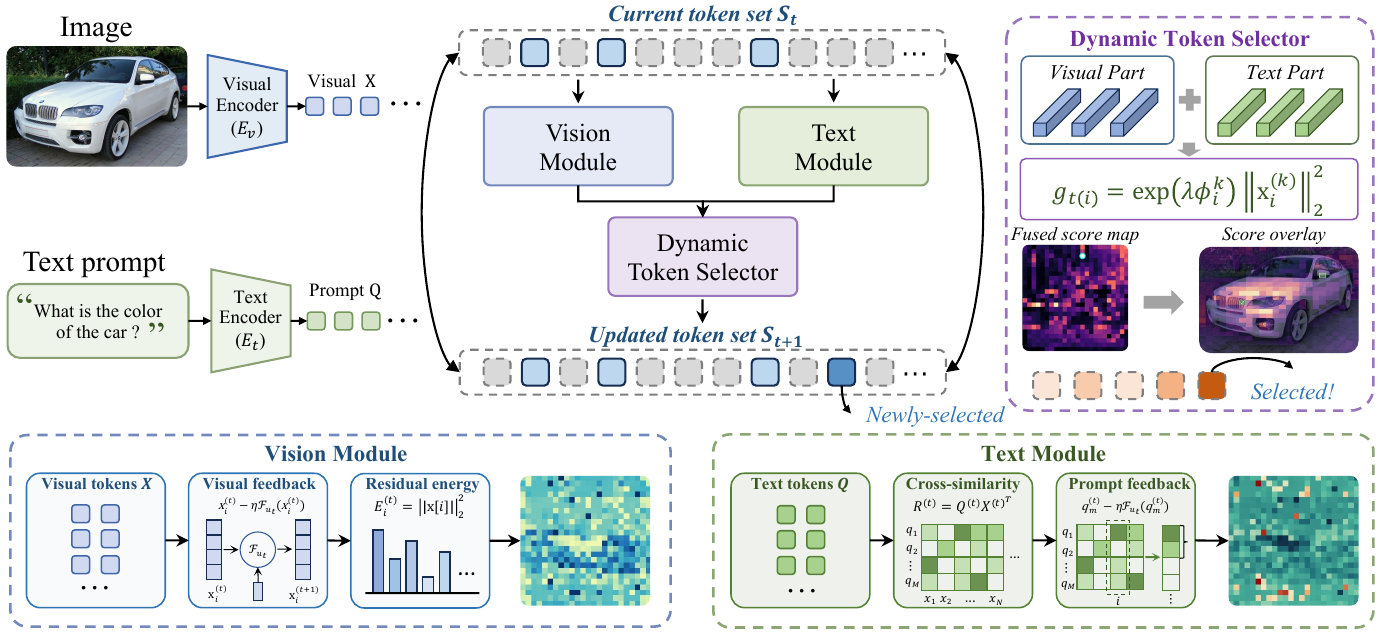}
    \caption{\textbf{Overview of DIVE}. Static top-$K$ pruning fixes all candidate scores before constructing the retained subset. DIVE maintains a visual-text residual state and repeats three operations: select the token with the largest residual-conditioned score, update the state to discount the evidence explained by that token, and re-evaluate the remaining candidates. The selected tokens are finally restored to their original order and passed to the frozen VLM.}
    \label{fig:main_overview}
\end{figure*}

In this section, we present DIVE for efficient VLM inference. To reduce the visual sequence processed by the LLM, DIVE retains a subset of visual tokens $\{\mathbf{x}_s\}_{s\in\mathcal{S}}$, where $\mathcal{N}=\{1,\ldots,N\}$ denotes the full index set, $\mathcal{S}\subseteq\mathcal{N}$, and $|\mathcal{S}|=K<N$. Rather than selecting $\mathcal{S}$ from a fixed ranking, DIVE constructs it iteratively through a \textit{select--update--re-evaluate} process, so that each selection updates the scores of the remaining candidates. Figure~\ref{fig:main_overview} illustrates the framework.

\subsection{Inputs and Residual States}
\label{sec:inputs_residual_states}
At the pruning layer, DIVE extracts visual and textual representations from the same multimodal sequence. We denote the visual representations by $\mathbf{X}=[\mathbf{x}_1,\ldots,\mathbf{x}_N]^\top\in\mathbb{R}^{N\times d}$ and the representations at the valid textual prompt positions by $\mathbf{Q}=[\mathbf{q}_1,\ldots,\mathbf{q}_M]^\top\in\mathbb{R}^{M\times d}$, excluding visual and padding positions from $\mathbf{Q}$. Here, $\mathbf{x}_i$ denotes visual candidate $i$, and $\mathbf{q}_m$ denotes prompt token $m$. Because $\mathbf{X}$ and $\mathbf{Q}$ are extracted from the same multimodal sequence, they lie in a \textbf{shared representation} space and can be directly compared through inner products.

Before selection, DIVE applies row-wise $\ell_2$ normalization to $\mathbf{X}$ and $\mathbf{Q}$. At iteration $t$, we denote their evolving working states by $\mathbf{X}^{(t)}$ and $\mathbf{Q}^{(t)}$, with $\mathbf{x}_i^{(t)}$ and $\mathbf{q}_m^{(t)}$ denoting the corresponding rows. The visual state $\mathbf{x}_i^{(t)}$ tracks the evidence in candidate $i$ that remains after previous selections, while the query state $\mathbf{q}_m^{(t)}$ tracks the prompt direction that remains to be covered. These states are updated only to guide selection; the original visual representations remain unchanged and are used for subsequent VLM inference.

\begin{algorithm}[t]
\caption{DIVE selection procedure}
\label{alg:dive_loop}
\textbf{Input}: Visual representations $\mathbf{X}$, textual prompt representations $\mathbf{Q}$, budget $K$, threshold $\epsilon$\\
\textbf{Output}: Retained visual tokens $\widetilde{\mathbf{X}}$
\begin{algorithmic}[1]
\STATE Normalize $\mathbf{X}$ and $\mathbf{Q}$ to initialize
       $\mathbf{X}^{(0)}$ and $\mathbf{Q}^{(0)}$; $\mathcal{S}_0\gets\varnothing$
\FOR{$t=0,\ldots,K-1$}
    \STATE $\mathcal{C}_t\gets\mathcal{N}\setminus\mathcal{S}_t$
    \STATE Compute $g_t(i)$ for all $i\in\mathcal{C}_t$ by
           Eq.~\eqref{eq:dive_score}
    \IF{$\max_{i\in\mathcal{C}_t}\|\mathbf{x}_i^{(t)}\|_2\leq\epsilon$}
        \STATE Append the first $K-t$ indices of $\mathcal{C}_t$ in input order to $\mathcal{S}_t$; \textbf{break}
    \ENDIF
    \STATE $s_t\gets\operatorname*{arg\,max}_{i\in\mathcal{C}_t}g_t(i)$
    \STATE $\mathcal{S}_{t+1}\gets\mathcal{S}_t\cup\{s_t\}$
    \STATE $\mathbf{u}_t\gets\mathbf{x}_{s_t}^{(t)}/\|\mathbf{x}_{s_t}^{(t)}\|_2$
    \STATE Update $\mathbf{X}^{(t+1)}$ and $\mathbf{Q}^{(t+1)}$ by
           Eq.~\eqref{eq:residual_updates}
\ENDFOR
\STATE $\widetilde{\mathbf{X}}\gets$ selected tokens in their input order
\RETURN $\widetilde{\mathbf{X}}$
\end{algorithmic}
\end{algorithm}

\subsection{Prompt-Conditioned Candidate Scoring}
\label{sec:prompt_conditioned_residual_scoring}
At iteration $t$, a useful candidate should provide visual evidence that remains uncovered while also being relevant to the unresolved prompt. Residual energy alone may favor visually complementary but prompt-irrelevant candidates, whereas prompt alignment alone may repeatedly favor candidates that match the same query semantics. DIVE therefore combines the current visual residual energy with a prompt-alignment weight, prioritizing candidates that provide uncovered evidence relevant to the unresolved prompt. The score of visual candidate $i$ is defined as
\begin{equation}
    g_t(i)
    =
    \exp\!\left(\lambda\phi_i^{(t)}\right)
    \left\|\mathbf{x}_i^{(t)}\right\|_2^2.
\label{eq:dive_score}
\end{equation}
The residual-energy term $\|\mathbf{x}_i^{(t)}\|_2^2$ measures how much visual evidence remains in candidate $i$ after previous selections. The prompt-alignment weight $\phi_i^{(t)}$ measures how strongly the candidate aligns with the current unresolved query state. At each iteration, DIVE computes the positive candidate-normalized alignments between the candidate and all query states, averages the three largest values, and normalizes the result across the remaining candidates. Its explicit computation is provided in Appendix~\ref{app:query_alignment_function}. The coefficient $\lambda$ controls the strength of prompt conditioning.

Because both the visual and query residual states evolve after every selection, the score $g_t(i)$ is re-evaluated at each iteration. Consequently, candidates that overlap with previously selected evidence lose residual energy, whereas candidates carrying distinct and prompt-relevant evidence remain competitive.

\subsection{Selection-Time Feedback Updates}
\label{sec:selection_time_feedback}

After selecting $s_t$, DIVE uses its current visual residual $\mathbf{x}_{s_t}^{(t)}$ as selection-time feedback and normalizes it to obtain the direction $\mathbf{u}_t$. This direction represents the visual evidence newly covered by the selected token. To propagate this feedback, DIVE discounts components that are positively aligned with $\mathbf{u}_t$. For a residual vector $\mathbf{z}$, we define the feedback component induced by $\mathbf{u}_t$ as
\begin{equation}
    \mathcal{F}_{\mathbf{u}_t}(\mathbf{z})
    =
    \left[
    \left\langle\mathbf{z},\mathbf{u}_t\right\rangle
    \right]_{+}\mathbf{u}_t,
\label{eq:feedback_component}
\end{equation}
where $[a]_{+}=\max(a,0)$. The operator $\mathcal{F}_{\mathbf{u}_t}$ extracts only the component positively aligned with the newly selected evidence. Because the visual and query residuals lie in the same representation space, DIVE applies the same \textbf{feedback direction} to both states:

\begin{table*}[!t]
    \centering
    \begingroup
\normalsize
\setlength{\tabcolsep}{2.3mm}
\colorlet{mainresgroup}{black!10}
\colorlet{mainresours}{orange!12}
\arrayrulecolor{black}
\newcommand{\twoLineCell}[2]{\begin{tabular}[c]{@{}c@{}}#1\\#2\end{tabular}}
\begin{tabular}{lc||cccccccc||c}
\toprule
\textbf{Method} & \textbf{Venue}
& \twoLineCell{\textbf{GQA}}{Acc. $\uparrow$}
& \twoLineCell{\textbf{MME}}{P+C $\uparrow$}
& \twoLineCell{\textbf{POPE}}{F1 $\uparrow$}
& \twoLineCell{\textbf{SQA}}{Acc. $\uparrow$}
& \twoLineCell{\textbf{VQA$^{\mathrm{t}}$}}{Acc. $\uparrow$}
& \twoLineCell{\textbf{OCR}}{Score $\uparrow$}
& \twoLineCell{\textbf{VQA$^{\mathrm{v2}}$}}{Acc. $\uparrow$}
& \twoLineCell{\textbf{VizWiz}}{Acc. $\uparrow$}
& \twoLineCell{\textbf{Avg.}}{$\uparrow$} \\
\midrule
\rowcolor{mainresgroup}
\multicolumn{2}{l}{\textbf{\textit{Full-token baseline}}} & \multicolumn{8}{c}{\textbf{\textit{576 visual tokens}}} & \\
Vanilla & -- & 61.9 & 1862 & 85.9 & 69.5 & 58.2 & 297 & 78.5 & 50.0 & 100.00\% \\
\midrule
\rowcolor{mainresgroup}
\multicolumn{2}{l}{\textbf{\textit{Retain 192 visual tokens}}} & \multicolumn{8}{c}{\textbf{\textit{66.7\% token reduction}}} & \\
FastV & ECCV24 & 52.7 & 1612 & 64.8 & 67.3 & 52.5 & 291 & 67.1 & 50.8 & 89.91\% \\
PDrop & CVPR25 & 57.1 & 1766 & 82.3 & 68.8 & 56.1 & 290 & 75.1 & 51.1 & 96.72\% \\
SparseVLM & ICML25 & 57.6 & 1721 & 83.6 & \textbf{69.1} & 56.1 & 292 & 75.6 & 50.5 & 96.78\% \\
VisionZip & CVPR25 & \underline{59.3} & \underline{1783} & 85.3 & \underline{68.9} & \underline{57.3} & \textbf{308} & \underline{76.8} & \underline{51.5} & \underline{99.12\%} \\
PruneSID & ICLR26 & 59.2 & 1769 & \textbf{87.1} & 68.8 & 56.7 & 288 & \underline{76.8} & \underline{51.5} & 98.28\% \\
\textbf{DIVE} & -- & \textbf{61.4} & \textbf{1826} & \underline{86.4} & \textbf{69.1} & \textbf{58.0} & \underline{304} & \textbf{78.1} & \textbf{52.1} & \textbf{100.37\%} \\
\midrule
\rowcolor{mainresgroup}
\multicolumn{2}{l}{\textbf{\textit{Retain 128 visual tokens}}} & \multicolumn{8}{c}{\textbf{\textit{77.8\% token reduction}}} & \\
FastV & ECCV24 & 49.6 & 1490 & 59.6 & 60.2 & 50.6 & 285 & 61.8 & 51.3 & 85.05\% \\
PDrop & CVPR25 & 56.0 & 1644 & 82.3 & \underline{68.3} & 55.1 & 287 & 72.9 & 51.0 & 94.88\% \\
SparseVLM & ICML25 & 56.0 & 1696 & 80.5 & 67.1 & 54.9 & 280 & 73.8 & 51.4 & 94.65\% \\
VisionZip & CVPR25 & 57.6 & \underline{1762} & 83.2 & \textbf{68.9} & \underline{56.8} & \underline{300} & \underline{75.6} & 51.8 & \underline{97.77\%} \\
PruneSID & ICLR26 & \underline{58.0} & 1739 & \underline{85.6} & 68.2 & 54.7 & 284 & 75.3 & \textbf{52.3} & 96.88\% \\
\textbf{DIVE} & -- & \textbf{60.8} & \textbf{1828} & \textbf{86.2} & \textbf{68.9} & \textbf{57.5} & \textbf{302} & \textbf{77.6} & \underline{52.2} & \textbf{99.95\%} \\
\midrule
\rowcolor{mainresgroup}
\multicolumn{2}{l}{\textbf{\textit{Retain 64 visual tokens}}} & \multicolumn{8}{c}{\textbf{\textit{88.9\% token reduction}}} & \\
FastV & ECCV24 & 46.1 & 1256 & 48.0 & 51.1 & 47.8 & 245 & 55.0 & 50.8 & 75.95\% \\
PDrop & CVPR25 & 41.9 & 1092 & 55.9 & 68.6 & 45.9 & 250 & 69.2 & 50.7 & 80.34\% \\
SparseVLM & ICML25 & 52.7 & 1505 & 75.1 & 62.2 & 51.8 & 180 & 68.2 & 50.1 & 84.95\% \\
VisionZip & CVPR25 & 55.1 & \underline{1690} & 77.0 & \underline{69.0} & \underline{55.5} & \underline{283} & 72.4 & \textbf{52.9} & 94.67\% \\
PruneSID & ICLR26 & \underline{57.1} & 1688 & \underline{83.8} & 68.0 & 54.2 & 268 & \underline{73.7} & \textbf{52.9} & \underline{95.17\%} \\
\textbf{DIVE} & -- & \textbf{59.7} & \textbf{1761} & \textbf{85.7} & \textbf{69.2} & \textbf{56.4} & \textbf{285} & \textbf{76.2} & \underline{52.8} & \textbf{98.21\%} \\
\bottomrule
\end{tabular}
\arrayrulecolor{black}
\endgroup

    \caption{Main results on LLaVA-1.5-7B under different token budgets. OCR denotes OCRBench, and Avg. denotes the relative average performance compared with the vanilla Full-token baseline. The best and second-best results at each budget are bolded and underlined, respectively.}
    \label{tab:main_results}
\end{table*}

\begin{equation}
    \begin{aligned}
        \mathbf{x}_{i}^{(t+1)} & = \mathbf{x}_{i}^{(t)} - \eta \mathcal{F}_{\mathbf{u}_t}\left(\mathbf{x}_{i}^{(t)}\right),
        \\
        \mathbf{q}_{m}^{(t+1)} & = \mathbf{q}_{m}^{(t)} - \eta \mathcal{F}_{\mathbf{u}_t}\left(\mathbf{q}_{m}^{(t)}\right).
    \end{aligned}
\label{eq:residual_updates}
\end{equation}
The coefficient $\eta$ controls the feedback strength. The visual update suppresses candidate evidence already covered by the selected token, while the query update discounts prompt directions that have been partially addressed. Because $\mathcal{F}_{\mathbf{u}_t}$ is zero for orthogonal and negatively aligned components, the update preserves evidence that is distinct from the current selection. The resulting change in residual energy is derived in Appendix~\ref{app:one_sided_residual_update}. We use one fixed configuration throughout all experiments: top-three alignment pooling, $\lambda=1.5$, and $\eta=0.8$. These settings remain unchanged across backbones, benchmarks, and token budgets.

\subsection{Dynamic Iterative Selection}
\label{sec:iterative_state_update}

Starting from $\mathcal{S}_0=\varnothing$, DIVE constructs the retained set one token at a time. At iteration $t$, DIVE scores every unselected candidate using Eq.~\eqref{eq:dive_score} and selects the highest-scoring candidate, denoted by $s_t$. It then adds the selected index to the retained set as $\mathcal{S}_{t+1}=\mathcal{S}_t\cup\{s_t\}$. The residual $\mathbf{x}_{s_t}^{(t)}$ provides the feedback direction, and Eq.~\eqref{eq:residual_updates} updates the visual and query states before the remaining candidates are re-evaluated.

Because the candidate scores depend on the evolving residual states, their relative ordering can change after each selection. Candidates overlapping with the retained evidence lose residual energy, while candidates carrying complementary and prompt-relevant evidence remain competitive. DIVE repeats this \textit{select--update--re-evaluate} process until $K$ tokens have been selected. It then restores the selected indices to their original spatial-temporal order and passes the corresponding original visual representations, rather than their residual states, to the frozen VLM. Algorithm~\ref{alg:dive_loop} summarizes the complete procedure, with initialization details and computational complexity provided in Appendix~\ref{app:representation_initialization} and Appendix~\ref{app:algorithm_complexity}.

\section{Experiments}
\label{sec:experiments}

\begin{table*}[!t]
    \centering
    \begingroup
\small
\setlength{\tabcolsep}{1.2mm}
\colorlet{summaryours}{orange!12}
\arrayrulecolor{black}
\begin{tabular}{l|ccc|ccc|ccc|ccc}
\toprule
\raisebox{-7pt}[0pt][0pt]{\textbf{\normalsize{Method}}}
& \multicolumn{3}{c}{\textbf{LLaVA-1.5-7B}}
& \multicolumn{3}{c}{\textbf{LLaVA-1.5-13B}}
& \multicolumn{3}{c}{\textbf{LLaVA-NeXT-7B}}
& \multicolumn{3}{c}{\textbf{LLaVA-NeXT-13B}} \\
\cmidrule(lr){2-4}\cmidrule(lr){5-7}\cmidrule(lr){8-10}\cmidrule(l){11-13}
& \multicolumn{3}{c}{\textit{576 tokens}}
& \multicolumn{3}{c}{\textit{576 tokens}}
& \multicolumn{3}{c}{\textit{Upper(Up.) 2880 tokens}}
& \multicolumn{3}{c}{\textit{Upper(Up.) 2880 tokens}} \\
\cmidrule(lr){1-1}\cmidrule(lr){2-4}\cmidrule(lr){5-7}\cmidrule(lr){8-10}\cmidrule(l){11-13}
\textbf{Compress Ratio}
& $\downarrow$67\% & $\downarrow$78\% & $\downarrow$89\%
& $\downarrow$67\% & $\downarrow$78\% & $\downarrow$89\%
& $\downarrow$78\% & $\downarrow$89\% & $\downarrow$94\%
& $\downarrow$78\% & $\downarrow$89\% & $\downarrow$94\% \\
\cmidrule(lr){1-1}\cmidrule(lr){2-2}\cmidrule(lr){3-3}\cmidrule(lr){4-4}
\cmidrule(lr){5-5}\cmidrule(lr){6-6}\cmidrule(lr){7-7}
\cmidrule(lr){8-8}\cmidrule(lr){9-9}\cmidrule(lr){10-10}
\cmidrule(lr){11-11}\cmidrule(lr){12-12}\cmidrule(lr){13-13}
\textbf{Remain Token}
& 192 & 128 & 64
& 192 & 128 & 64
& Up. 640 & Up. 320 & Up. 160
& Up. 640 & Up. 320 & Up. 160 \\
\midrule
FastV [ECCV24]
& 89.9\% & 85.0\% & 76.0\% & 98.0\% & 96.0\% & 91.0\% & 89.4\% & 84.5\% & 75.1\% & 91.4\% & 86.7\% & 78.4\% \\
VisionZip [CVPR25]
& 99.1\% & 97.8\% & 94.7\% & 97.7\% & 95.8\% & 93.2\% & 87.7\% & 89.6\% & 84.4\% & 87.4\% & 86.5\% & 84.1\% \\
PruneSID [ICLR26]
& 98.3\% & 96.9\% & 95.2\% & 92.5\% & 90.6\% & 90.4\% & 92.5\% & 89.7\% & 82.8\% & 93.2\% & 89.0\% & 78.6\% \\
\midrule
\textbf{DIVE [OURS]}
& \textbf{100.3\%} & \textbf{99.9\%} & \textbf{98.2\%}
& \textbf{99.6\%} & \textbf{99.1\%} & \textbf{97.4\%}
& \textbf{94.2\%} & \textbf{90.9\%} & \textbf{86.7\%}
& \textbf{94.4\%} & \textbf{91.2\%} & \textbf{87.2\%} \\
\bottomrule
\end{tabular}
\arrayrulecolor{black}
\endgroup

    \caption{Comparison across LLaVA-1.5 and LLaVA-NeXT backbones under different retained-token budgets. }
    \label{tab:cross_backbone_summary}
\end{table*}

\subsection{Experiment Setting}
\noindent\textbf{Datasets and Evaluation.}
We evaluate DIVE on ten LMMS-EVAL~\cite{lmms-eval} benchmarks across three task categories:
1) \textit{Visual Question Answering (VQA)}: GQA~\cite{GQA}, ScienceQA~\cite{ScienceQA}, VQAv2~\cite{VQAv2}, VizWiz~\cite{VizWiz}, and TextVQA~\cite{TextVQA};
2) \textit{Visual Perception}: MME~\cite{MME}, POPE~\cite{POPE}, and OCRBench~\cite{OCRBench}; and
3) \textit{Video Understanding}: MVBench~\cite{MVBench} and VideoMME~\cite{VideoMME}.
Detailed descriptions and evaluation protocols are provided in Appendix~\ref{app:exp_setup}.

\noindent\textbf{Compared Methods.} We compare DIVE with representative visual-token compression methods from two categories:  
\textit{In-LLM Token Pruning} includes FastV~\cite{FastV}, SparseVLM~\cite{SparseVLM}, and PDrop~\cite{Pdrop}, which prune tokens during LLM inference using internal attention signals;  
and \textit{Pre-LLM Token Compression} includes VisionZip~\cite{visionzip}, DART~\cite{dart}, and PruneSID~\cite{PruneSID}, which reduce visual tokens before LLM inference based on importance, redundancy, or semantic cues.  
Detailed descriptions and implementation settings are provided in Appendix~\ref{app:exp_setup}.

\begin{table}[!t]
    \centering
    {\small
    \begingroup
\setlength{\tabcolsep}{1.6mm}
\colorlet{singlegrp}{black!10}
\colorlet{singleours}{orange!12}
\arrayrulecolor{black}
\begin{tabular}{@{}lccccc@{}}
\toprule
\textbf{Methods} & \textbf{SQA} & \textbf{POPE} & \textbf{MME} & \textbf{GQA} & \textbf{Avg.} \\
\midrule
\rowcolor{singlegrp}
\multicolumn{6}{l}{\textbf{\textit{Full-token baseline (100\%)}}} \\
Vanilla & 85.5 & 87.8 & 2354 & 62.2 & 100.0\% \\
\midrule
\rowcolor{singlegrp}
\multicolumn{6}{c}{\textbf{\textit{Token reduction ($\downarrow$66.7\%)}}} \\
FastV [ECCV24] & \underline{82.2} & 84.8 & \underline{2280} & 58.8 & 96.0\% \\
DART [EMNLP25] & 80.2 & 85.2 & 2201 & 58.4 & 94.6\% \\
VisionZip [CVPR25] & \underline{82.2} & \underline{85.5} & 2275 & \underline{59.5} & \underline{96.5\%} \\
\textbf{DIVE [OURS]} & \textbf{82.4} & \textbf{87.9} & \textbf{2306} & \textbf{61.3} & \textbf{98.3\%} \\
\midrule
\rowcolor{singlegrp}
\multicolumn{6}{c}{\textbf{\textit{Token reduction ($\downarrow$77.8\%)}}} \\
FastV [ECCV24] & \underline{80.5} & 81.4 & 2208 & 56.4 & 92.8\% \\
DART [EMNLP25] & 78.6 & 82.8 & 2094 & 55.8 & 91.2\% \\
VisionZip [CVPR25] & 79.9 & \underline{83.5} & \underline{2216} & \underline{57.3} & \underline{93.7\%} \\
\textbf{DIVE [OURS]} & \textbf{82.5} & \textbf{88.1} & \textbf{2268} & \textbf{60.7} & \textbf{97.7\%} \\
\bottomrule
\end{tabular}
\arrayrulecolor{black}
\endgroup
}
    \caption{Comparative Experiments on Qwen2-VL-7B.}
    \label{tab:qwen2vl_image}
\end{table}

\subsection{Main Comparisons}
\noindent\textbf{Image Understanding.} LLaVA-1.5-7B provides a controlled setting for evaluating visual-token pruning, with 576 visual tokens per image. As shown in Table~\ref{tab:main_results}, DIVE retains 100.3/99.9/98.2\% of the uncompressed model's average performance with 192/128/64 tokens, respectively. It achieves the highest average at all three budgets, outperforming its strongest competitors, VisionZip and PruneSID. At the task level, DIVE leads on GQA, MME, TextVQA, and VQAv2 across all three budgets and, with only 64 tokens, ranks first on seven of the eight benchmarks. 

\noindent\textbf{Cross-Backbone Results.} Starting from LLaVA-1.5-7B setting, we first scale the model to LLaVA-1.5-13B while retaining the same 576-token visual representation. As summarized in Table~\ref{tab:cross_backbone_summary}, DIVE remains the best-performing method at all three budgets and preserves 97.4\% of the uncompressed model's average performance with only 64 tokens. LLaVA-NeXT introduces a larger change by replacing the fixed-resolution representation with a high-resolution visual pipeline containing 2880 tokens. Across both its 7B and 13B variants, DIVE again achieves the highest relative average at every evaluated budget. We further evaluate Qwen2-VL, which differs from the LLaVA series in both model architecture and its flexible-resolution, variable-length visual representation. As shown in Table~\ref{tab:qwen2vl_image}, DIVE preserves 98.3/97.7\% of the uncompressed performance under 66.7/77.8\% token reduction, respectively, and ranks first on all four benchmarks at both ratios. Maintaining the performance lead through these progressively different settings demonstrates DIVE's strong effectiveness across diverse VLM architectures.

\begin{table}[!t]
    \centering
    {\small
    \begingroup
\setlength{\tabcolsep}{0.3mm}
\colorlet{singlegrp}{black!10}
\colorlet{singleours}{orange!12}
\arrayrulecolor{black}
\begin{tabular}{@{}lcccccc@{}}
\toprule
\textbf{Methods} & \textbf{MVB} & \multicolumn{4}{c}{\textbf{VideoMME}} & \textbf{Avg.} \\
\cmidrule(lr){3-6}
 &  & Overall & Short & Medium & Long &  \\
\midrule
\rowcolor{singlegrp}
\multicolumn{7}{l}{\textbf{\textit{Full-token baseline(100\%)}}} \\
Vanilla & 58.3 & 58.4 & 69.9 & 56.7 & 48.8 & 100.0\% \\
\midrule
\rowcolor{singlegrp}
\multicolumn{7}{c}{\textbf{\textit{Average retention ratio = 25\%}}} \\
FastV [ECCV24] & 54.9 & \underline{54.8} & 62.9 & \underline{54.4} & \underline{47.1} & \underline{94.1\%} \\
SparseVLM [ICML25] & 53.7 & 52.6 & 61.4 & 51.1 & 45.3 & 90.6\% \\
PDrop [CVPR25] & \underline{55.6} & 54.3 & \underline{65.6} & 52.9 & 44.4 & 93.3\% \\
\textbf{DIVE [OURS]} & \textbf{57.1} & \textbf{58.1} & \textbf{70.8} & \textbf{56.0} & \textbf{47.6} & \textbf{99.0\%} \\
\midrule
\rowcolor{singlegrp}
\multicolumn{7}{c}{\textbf{\textit{Average retention ratio = 15\%}}} \\
FastV [ECCV24] & 52.5 & \underline{53.4} & 60.6 & \underline{53.0} & \underline{46.7} & \underline{91.5\%} \\
SparseVLM [ICML25] & 50.5 & 51.1 & 57.4 & 50.9 & 45.1 & 87.7\% \\
PDrop [CVPR25] & \underline{53.5} & 52.7 & \underline{63.3} & 50.7 & 44.1 & 90.5\% \\
\textbf{DIVE [OURS]} & \textbf{57.1} & \textbf{57.5} & \textbf{67.8} & \textbf{56.7} & \textbf{48.0} & \textbf{98.4\%} \\
\bottomrule
\end{tabular}
\arrayrulecolor{black}
\endgroup
}
    \caption{Comparative Experiments on LLaVA-OV-7B in video understanding benchmarks.}
    \label{tab:llava_ov_video}
\end{table}

\noindent\textbf{Video Understanding.} Video compression is particularly challenging because relevant evidence may be distributed across both spatial regions and temporal frames. We evaluate DIVE on two video-capable architectures, LLaVA-OV-7B and Qwen2-VL-7B. As shown in Table~\ref{tab:llava_ov_video} and Appendix Table~\ref{tab:qwen2vl_video}, DIVE preserves at least 98.0\% of the uncompressed performance while retaining only 15\% of the visual tokens, outperforming all competing methods. Its advantage is consistent across MVBench and the short/medium/long duration splits of VideoMME, showing that DIVE remains effective across different types and scales of video understanding.

\begin{table*}[!t]
    \centering
    \begingroup
\setlength{\tabcolsep}{1mm}
\colorlet{singlegrp}{black!10}
\newcommand{\effTwoLineCell}[2]{\begin{tabular}[c]{@{}c@{}}#1\\#2\end{tabular}}
\begin{tabular}{lcccccccc}
\toprule
\textbf{Method} & \textbf{Tokens $\downarrow$}
& \effTwoLineCell{\textbf{Total $\downarrow$}}{\textbf{(Min:sec)}}
& \effTwoLineCell{\textbf{Prefill $\downarrow$}}{\textbf{(Min:sec)}}
& \effTwoLineCell{\textbf{GPU Mem $\downarrow$}}{\textbf{(GB)}}
& \effTwoLineCell{\textbf{KV Cache $\downarrow$}}{\textbf{(MB)}}
& \effTwoLineCell{\textbf{POPE $\uparrow$}}{\textbf{(F1-score)}}
& \multicolumn{2}{c}{\begin{tabular}[c]{@{}cc@{}}\multicolumn{2}{c}{\textbf{Speedup $\uparrow$}}\\\textbf{(Total)} & \textbf{(Prefilling)}\end{tabular}} \\
\midrule
\rowcolor{singlegrp}
Vanilla & 576 & 32:11 & 23:12 & 14.25 & 318.85 & 85.9 & 1.00$\times$ & 1.00$\times$ \\
VisionZip [CVPR25]& 64 & 20:32 & 9:05 & 13.59 & \textbf{62.85} & 77.0 & 1.57$\times$ & 2.55$\times$ \\
PruneSID [ICLR26]& 64 & 22:19 & \textbf{9:01} & 13.63 & \textbf{62.85} & 83.8 & 1.44$\times$ & \textbf{2.57$\times$} \\
SparseVLM [ICML25]& 64 & 28:40 & 17:47 & 18.34 & 134.10 & 75.1 & 1.12$\times$ & 1.30$\times$ \\
\textbf{DIVE [OURS]} & 64 & \textbf{19:07} & 10:15 & \textbf{13.48} & 78.85 & \textbf{85.7} & \textbf{1.68$\times$} & 2.26$\times$ \\

\bottomrule
\end{tabular}
\arrayrulecolor{black}
\endgroup

    \caption{Efficiency comparison on LLaVA-1.5-7B and 64-token compressed variants.}
    \label{tab:efficiency}
\end{table*}

\begin{figure*}[t]
    \centering
    \includegraphics[width=1\textwidth]{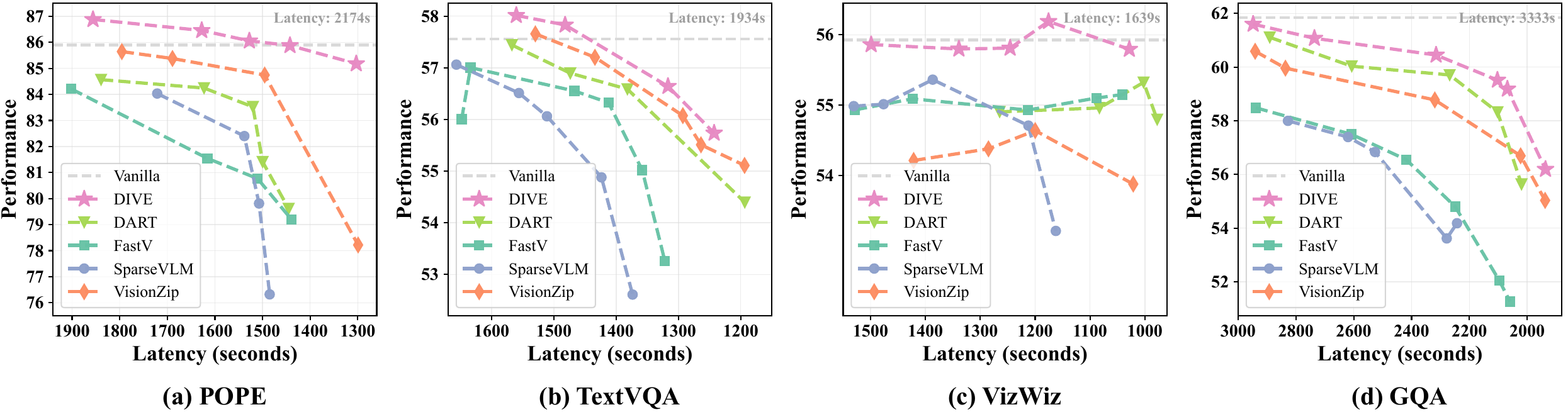}
    \caption{\textbf{End-to-end performance--latency trade-offs} across POPE, TextVQA, VizWiz, and GQA on LLaVA-1.5-7B. Latency denotes full-benchmark wall-clock time, including inference, prediction saving, and metric computation. All performance values are reported as accuracy (\%).}
    \label{fig:performance_latency_tradeoff}
\end{figure*}

Taken together, these results demonstrate that DIVE consistently delivers leading performance across image and video tasks, model scales, and VLM architectures under substantial token compression. Its effectiveness across fixed-resolution images, high- and flexible-resolution inputs, and long spatiotemporal sequences highlights the broad applicability of iterative visual-evidence selection.

\subsection{Efficiency Analysis}
The reduction in the number of visual tokens only provides potential computational savings; for methods that require additional token selection, the critical question is whether these savings can translate into actual end-to-end acceleration after factoring in the selection overhead. To this end, we analyze the runtime efficiency under a fixed token budget and the performance variations under different runtime budgets.

\noindent\textbf{Computational Efficiency.}
Table~\ref{tab:efficiency} compares the practical inference efficiency of different methods on POPE with LLaVA-1.5-7B. By reducing the visual sequence from 576 to 64 tokens, DIVE decreases the prefill time from 23:12 to 10:15, yielding a $2.26\times$ speedup, and reduces the total runtime from 32:11 to 19:07, yielding a $1.68\times$ end-to-end speedup, with only a 0.2-point drop in F1. Although its prefill speedup is slightly lower than those of VisionZip and PruneSID, DIVE achieves both the highest F1 and the shortest total runtime. After subtracting prefill time, DIVE spends only 8:52 in post-prefill processing, compared with 11:27 for VisionZip, 13:18 for PruneSID, and 10:53 for SparseVLM. This comparison shows that prefill latency or KV-cache size alone does not fully characterize practical efficiency. By trading a moderate selection cost for better-preserved visual evidence, DIVE achieves the best overall performance--efficiency trade-off among the evaluated methods.

\noindent\textbf{Performance--Latency Trade-off.}
Figure~\ref{fig:performance_latency_tradeoff} evaluates whether the fixed-budget result extends across different runtime regimes. Across POPE, TextVQA, VizWiz, and GQA, DIVE consistently traces the upper envelope of the performance--latency curves, preserving stronger task performance over a broad range of measured runtimes. The difference becomes more pronounced as the latency budget decreases: competing methods often obtain additional speed through more aggressive compression, but their performance deteriorates rapidly, whereas DIVE exhibits a substantially more gradual degradation. This consistent trend shows that the advantage of DIVE is not tied to a particular token budget or an isolated operating point. More importantly, it indicates that practical efficiency depends not only on how much computation is removed, but also on how much useful visual evidence remains after compression. By allocating the retained-token budget to complementary and task-relevant evidence, DIVE converts reduced runtime into a more favorable performance--latency trade-off rather than merely pursuing the lowest possible latency.

\subsection{Ablation and Mechanism Validation}
\begin{figure*}[!t]
    \centering
    \includegraphics[width=\textwidth]{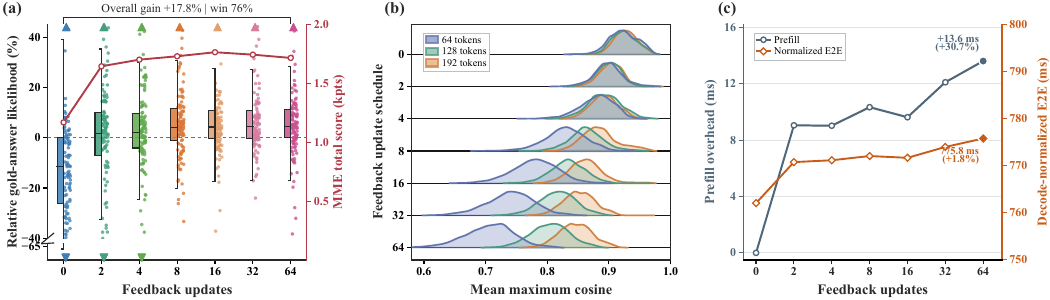}
    \caption{\textbf{Mechanism validation of set-dependent residual
    feedback.}
    (a) \textit{Correct-Answer Support:} Relative
    gold-answer likelihood under a 64-token budget. Boxes and dots show the likelihood distributions, red curve reports the score. (b) \textit{Retained-Set Redundancy:} Mean maximum cosine similarity across feedback schedules with 64/128/192 retained tokens. (c) \textit{Runtime Cost:} Paired mean prefill overhead and decode-normalized end-to-end latency across feedback schedules with 64 retained visual tokens and 32 generated tokens, measured over all MME samples.}
    \label{fig:dynamic_feedback_analysis}
\end{figure*}

The central novelty of DIVE lies in its \textbf{set-dependent residual feedback} mechanism: for each selection, DIVE follows a \textit{select--update--re-evaluate} process, allowing token utilities to evolve with the retained subset. We first examine the contributions of individual components in Table~\ref{tab:llava15_ablation}. We then validate the residual-feedback mechanism in Figure~\ref{fig:dynamic_feedback_analysis} through three complementary analyses: \textit{Correct-Answer Support}, \textit{Retained-Set Redundancy}, and \textit{Runtime Cost}.

\begin{table}[!t]
    \centering
    {\small
    \begingroup
\small
\setlength{\tabcolsep}{0.38mm}
\setlength{\overfullrule}{0pt}
\colorlet{singlegrp}{black!10}
\colorlet{singleours}{orange!12}
\arrayrulecolor{black}
\begin{tabular}{lcccccccc}
\toprule
\textbf{Method} & \textbf{GQA} & \textbf{MME} & \textbf{POPE} & \textbf{SQA} & \textbf{VQA$^{\mathrm{t}}$} & \textbf{OCR} & \textbf{Avg.} & \textbf{Thr.} \\
\midrule
\rowcolor{singlegrp}
\multicolumn{9}{l}{\textbf{\textit{Full-token baseline(100\%)}}} \\
Vanilla & 61.9 & 1862 & 85.9 & 69.5 & 58.2 & 297 & 100.00\% & 1.00$\times$ \\
\midrule
\rowcolor{singlegrp}
\multicolumn{9}{c}{\textbf{\textit{Retain 64 tokens ($\downarrow$88.9\%)}}} \\
w/o L-G & \underline{59.6} & 1741 & \underline{84.2} & 68.3 & \underline{56.3} & \underline{279} & \underline{96.13\%} & \underline{1.61$\times$} \\
w/o V-C & 58.8 & \underline{1758} & 83.8 & \underline{68.9} & 56.0 & 274 & 95.76\% & 1.58$\times$ \\
No update & 46.1 & 1169 & 43.5 & 65.1 & 44.5 & 41 & 61.97\% & \textbf{1.65$\times$} \\
\textbf{DIVE} & \textbf{59.7} & \textbf{1761} & \textbf{85.7} & \textbf{69.2} & \textbf{56.4} & \textbf{285} & \textbf{97.20\%} & 1.55$\times$ \\
\bottomrule
\end{tabular}
\arrayrulecolor{black}
\endgroup
}
    \caption{Ablation on LLaVA-1.5-7B with a 64-token budget. w/o L-G, w/o V-C, and No Update remove \textit{language guidance}, \textit{visual-centric scoring}, and \textit{iterative residual updates}, respectively. Avg. denotes average relative performance, and Thr. denotes throughput speedup. }
    \label{tab:llava15_ablation}
\end{table}

\noindent\textbf{Component Ablation.}
Table~\ref{tab:llava15_ablation} compares three ablations under the same 64-token setting: w/o L-G removes language guidance, while w/o V-C removes visual-centric scoring. No Update uses the same scoring function, initialization, and hyperparameters as DIVE, but freezes the candidate scores at initialization and selects the top 64 tokens from this one-shot ranking. This ablation tests whether DIVE's performance can be explained by its initial ranking alone, rather than serving as a competitive static pruning baseline. DIVE retains 97.20\% of the vanilla performance, and the substantial gap to No Update shows that set-dependent residual feedback is necessary within the DIVE formulation.

\paragraph{Correct-Answer Support.}
\label{par:correct_answer_support}
Under a 64-token budget, we vary the number of \textit{feedback updates} from zero (Frozen Score) to 64 (Full Update), with intermediate updates evenly distributed throughout selection. Figure~\ref{fig:dynamic_feedback_analysis}(a) reports the teacher-forced gold-answer likelihood, normalized per sample by its geometric mean across schedules, together with the corresponding MME score. Higher likelihood indicates stronger correct-answer support. \textit{Even a few updates improve both measures}, while the gains diminish and largely stabilize beyond 16 updates. This trend shows that residual feedback progressively strengthens the retained evidence, with a moderate update frequency capturing most of the benefit.

\paragraph{Retained-Set Redundancy.}
Using the same feedback schedules, we evaluate token redundancy under 64/128/192 token budgets. Figure~\ref{fig:dynamic_feedback_analysis}(b) reports the \textit{mean maximum cosine similarity}, obtained by averaging each retained token's highest similarity to another token; lower values indicate less redundancy. The vertical axis denotes the feedback schedule, and the colors denote token budgets. \textit{Finer-grained feedback shifts the distributions toward lower similarity across all budgets}, with the largest reduction under the 64-token setting. This shows that residual feedback reduces within-set redundancy, particularly when the budget is constrained.

\paragraph{Runtime Cost.}
Under a 64-token budget and 32 generated tokens, we compare feedback schedules using prefill overhead and decode-normalized end-to-end latency. Relative to Frozen Score, Full Update increases mean prefill latency by 13.6 ms (30.7\%). However, the corresponding end-to-end latency increases by only \textbf{1.8\%}. Thus, the relative cost of iterative updating is largely amortized over the complete inference process. These results show that residual feedback introduces additional computation during prefill while having only a limited impact on overall latency.

\section{Conclusion}
\label{sec:conclusion}

DIVE reframes visual token pruning from \textit{static top-$k$ pruning} into a \textit{select--update--re-evaluate} strategy, where each selection changes the value of the remaining tokens. By suppressing evidence already covered, DIVE guides later selections toward complementary, task-relevant content and performs better than static selection. Results across the VLMs, token budgets, and image and video tasks show that this mechanism remains effective across diverse compression settings. These findings suggest that visual-token pruning should consider not only the importance of each token, but also the new evidence it adds to the retained set. DIVE thus offers a set-dependent perspective for designing visual token pruning in high-resolution images and long-video scenarios.

\bibliography{bib/aaai2027}

\appendix

\twocolumn[
\begin{center}
    {\LARGE\bfseries \papertitle\par}
    \medskip
    {\Large\bfseries Supplementary Material\par}
    \medskip
\end{center}
]

\noindent Appendix~A describes the algorithmic details of DIVE's
\textit{select--update--re-evaluate} procedure, along with an analysis of its
computational complexity. Appendix~B provides details of the experimental
setup, including model architectures, evaluation benchmarks, comparison
methods, and implementation details. Appendices~C and~D present additional
experimental and visualization results, respectively. Appendix~E discusses
the limitations of DIVE.

\section{Method Details}
\label{app:method_details}

\subsection{Inputs and Initialization}
\label{app:representation_initialization}

\noindent\textbf{Shared representation.}
Let $\mathbf{H}$ denote the hidden states of the complete multimodal sequence at the selection point. We use $\mathcal{I}_{\mathrm v}$ to index the visual positions and $\mathcal{I}_{\mathrm q}$ to index the valid textual prompt positions, excluding visual and padding positions. DIVE gathers the corresponding rows as
\begin{equation}
    \begin{aligned}
        \mathbf{X}
        &=\mathbf{H}_{\mathcal{I}_{\mathrm v},:}
        =[\mathbf{x}_1,\ldots,\mathbf{x}_N]^\top
        \in\mathbb{R}^{N\times d},\\
        \mathbf{Q}
        &=\mathbf{H}_{\mathcal{I}_{\mathrm q},:}
        =[\mathbf{q}_1,\ldots,\mathbf{q}_M]^\top
        \in\mathbb{R}^{M\times d}.
    \end{aligned}
    \label{eq:shared_representation_extraction}
\end{equation}
Because $\mathbf{X}$ and $\mathbf{Q}$ are row subsets of the same representation matrix, they share a coordinate space and can be compared through inner products.  After selection, DIVE restores the retained indices to this input order and passes the corresponding original rows of $\mathbf{X}$ to the frozen LLM.

\noindent\textbf{Initialization.}
DIVE initializes its working residuals by row-wise $\ell_2$ normalization for all visual candidates and prompt tokens, using a clamped denominator $\epsilon=10^{-6}$:
\begin{equation}
    \begin{aligned}
        \mathbf{x}_i^{(0)}
        &=
        \frac{\mathbf{x}_i}{\max(\|\mathbf{x}_i\|_2,\epsilon)},
        \quad i\in\{1,\ldots,N\},\\
        \mathbf{q}_m^{(0)}
        &=
        \frac{\mathbf{q}_m}{\max(\|\mathbf{q}_m\|_2,\epsilon)},
        \quad m\in\{1,\ldots,M\}.
    \end{aligned}
    \label{eq:residual_initialization}
\end{equation}
Their row-stacked states at selection step $t$ are denoted by $\mathbf{X}^{(t)}$ and $\mathbf{Q}^{(t)}$. These residual states guide selection only; the original representations in Eq.~\eqref{eq:shared_representation_extraction} remain unchanged.

\subsection{Prompt-Alignment Score}
\label{app:query_alignment_function}

At iteration $t$, let $\mathcal{C}_t=\{1,\ldots,N\}\setminus\mathcal{S}_t$ denote the unselected visual candidates, where $\mathcal{S}_t$ contains the previously selected indices. For each $i\in\mathcal{C}_t$ and query token $m$, DIVE computes the candidate-normalized rectified alignment
\begin{equation}
    \alpha_{mi}^{(t)}
    =
    \left[
    \frac{\langle\mathbf{q}_m^{(t)},\mathbf{x}_i^{(t)}\rangle}
    {\max(\|\mathbf{x}_i^{(t)}\|_2,\epsilon)}
    \right]_{+},
    \label{eq:rectified_query_candidate_alignment}
\end{equation}
where $[z]_{+}=\max(z,0)$. To prevent many weakly related prompt tokens from dominating the score, let $h\in\{1,\ldots,M\}$ denote the pooling size; by default, $h=\min(3,M)$. Let $\mathcal{T}_i^{(t)}$ contain the indices of the $h$ largest values of $\alpha_{mi}^{(t)}$. The prompt-alignment weight is then
\begin{equation}
    \begin{aligned}
    r_i^{(t)}
    &=
    \frac{1}{h}
    \sum_{m\in\mathcal{T}_i^{(t)}}\alpha_{mi}^{(t)},\\
    \phi_i^{(t)}
    &=
    \frac{r_i^{(t)}}
    {\displaystyle\max_{j\in\mathcal{C}_t}r_j^{(t)}+\epsilon}.
    \end{aligned}
\label{eq:phi_explicit}
\end{equation}
The top-$h$ mean retains the strongest prompt evidence while suppressing weak matches, and the second line rescales it over the current candidate set. The alignments are obtained from the maintained cross-similarity matrix $\mathbf{Q}^{(t)}{\mathbf{X}^{(t)}}^\top$. Query residuals are not re-normalized after feedback updates, so their norms continue to encode how much unresolved prompt evidence remains.

Using the current prompt-alignment weight, the score at iteration $t$ is

\begin{equation}
    g_t(i)=
    \exp\!\left(\lambda\phi_i^{(t)}\right)
    \|\mathbf{x}_i^{(t)}\|_2^2.
    \label{eq:implemented_dive_score}
\end{equation}

\noindent\textbf{Effect of prompt-conditioning strength.}
Since $0\leq\phi_i^{(t)}<1$, prompt conditioning adds the bounded term $\lambda\phi_i^{(t)}$ in log-score space: $\log g_t(i)=\log\|\mathbf{x}_i^{(t)}\|_2^2+\lambda\phi_i^{(t)}$. Thus, $\lambda$ controls the trade-off between residual visual energy and prompt alignment; setting $\lambda=0$ removes language guidance. For $\lambda=1.5$, the prompt multiplier is less than $e^{1.5}\approx 4.48$. The sensitivity to $\lambda$ is reported in Table~\ref{tab:parameter_sensitivity}.

\subsection{Residual-Energy Interpretation}
\label{app:residual_gain_approximation}
\label{app:one_sided_residual_update}

\noindent\textbf{Proposition 1 (Monotone one-sided residual update).}
Let $\mathbf{z}$ be a residual vector and let $\|\mathbf{u}\|_2=1$ be the feedback direction induced by a selected token. The one-sided update $\mathbf{z}^{+}=\mathbf{z}-\eta\mathcal{F}_{\mathbf{u}}(\mathbf{z})$, with $\mathcal{F}_{\mathbf{u}}$ defined in Eq.~\eqref{eq:feedback_component}, satisfies
\begin{equation}
    \|\mathbf{z}^{+}\|_2^2
    =\|\mathbf{z}\|_2^2
    -\eta(2-\eta)[\langle\mathbf{z},\mathbf{u}\rangle]_{+}^{2}.
    \label{eq:monotone_residual_energy}
\end{equation}
Consequently, for $0<\eta<2$, the update never increases residual energy. The decrease is positive only for residual components aligned with the selected direction; orthogonal and negatively aligned components remain unchanged.

\noindent\emph{Proof.}
Let $a=\langle\mathbf{z},\mathbf{u}\rangle$. The rectified projection gives
\begin{equation}
    \mathbf{z}^{+}
    =
    \begin{cases}
    \mathbf{z}, & a\leq 0,\\
    \mathbf{z}-\eta a\mathbf{u}, & a>0.
    \end{cases}
    \label{eq:rectified_update_cases}
\end{equation}
The first case leaves the residual energy unchanged. For $a>0$, expanding the squared norm and using $\|\mathbf{u}\|_2=1$ gives
\begin{equation}
    \begin{aligned}
        \|\mathbf{z}^{+}\|_2^2
        &=\|\mathbf{z}-\eta a\mathbf{u}\|_2^2\\
        &=\|\mathbf{z}\|_2^2
        -2\eta a\langle\mathbf{z},\mathbf{u}\rangle
        +\eta^2a^2\|\mathbf{u}\|_2^2\\
        &=\|\mathbf{z}\|_2^2-\eta(2-\eta)a^2.
    \end{aligned}
    \label{eq:positive_alignment_energy_expansion}
\end{equation}
Combining the two cases proves Eq.~\eqref{eq:monotone_residual_energy}.

\noindent\textbf{Choice of update strength.}
The one-step decrease factor satisfies
\begin{equation}
    \eta(2-\eta)=1-(\eta-1)^2\leq 1,
    \label{eq:update_strength_bound}
\end{equation}
and is therefore maximized at $\eta=1$, which completely removes the positively aligned component in one update. However, positive alignment between visual tokens may indicate partial overlap rather than complete redundancy; we therefore use $\eta=0.8$ to conservatively discount shared evidence while retaining potentially useful aligned content. The sensitivity to $\eta$ is reported in Table~\ref{tab:parameter_sensitivity}.

\noindent\textbf{Candidate-wise decrease.}
At step $t$, selecting candidate $i\in\mathcal{C}_t$ with $\|\mathbf{x}_i^{(t)}\|_2>\epsilon$ induces the exact residual-energy decrease
\begin{equation}
    \begin{aligned}
        \Delta_{t,\mathrm v}(i)
        &=\sum_{j\in\mathcal{C}_t}
        \left(\|\mathbf{x}_j^{(t)}\|_2^2
        -\|\widetilde{\mathbf{x}}_{j\mid i}^{(t+1)}\|_2^2\right)\\
        &=\eta(2-\eta)
        \sum_{j\in\mathcal{C}_t}
        \bigl[\langle\mathbf{x}_j^{(t)},\mathbf{u}_i^{(t)}\rangle\bigr]_+^2,
    \end{aligned}
    \label{eq:candidate_energy_decrease}
\end{equation}
where $\widetilde{\mathbf{x}}_{j\mid i}^{(t+1)}$ denotes the residual after one
update along $\mathbf{u}_i^{(t)}=\mathbf{x}_i^{(t)}/\|\mathbf{x}_i^{(t)}\|_2$.
Its self-term is $\eta(2-\eta)\|\mathbf{x}_i^{(t)}\|_2^2$, so ranking by residual
norm ranks this contribution exactly. DIVE modulates this self-term with prompt
alignment, while iterative residual updates suppress overlapping candidates.
This residual-energy surrogate does not imply global set-function optimization
or an approximation guarantee for the downstream objective.

\noindent\textbf{Degenerate residuals.}
The analysis assumes that the selected visual residual has norm greater than
$\epsilon$. If all unselected visual residuals have norm at most $\epsilon$,
DIVE skips further feedback updates and completes the remaining selections in
ascending original-index order. This rule makes the zero-residual case
deterministic without affecting the nondegenerate procedure.

\subsection{Computational Cost}
\label{app:algorithm_complexity}

Let $N$, $M$, $K$, and $d$ denote the numbers of visual tokens, query tokens, retained visual tokens, and the embedding dimension. DIVE first forms the cross-similarity matrix in $O(MNd)$ time. At each of the $K$ selection steps, updating the visual and query residuals costs $O((N+M)d)$, while updating or scanning the maintained cross-similarity scores costs $O(MN)$. The total selection cost is therefore
\begin{equation}
    O\!\left(MNd+K((N+M)d+MN)\right),
    \label{eq:dive_complexity}
\end{equation}
rather than recomputing a full $M\times N$ matrix product at every step.

After fixing the configuration used in the main experiments, we further examined its sensitivity under the LLaVA-1.5-7B 64-token setting.

\begin{table*}[!t]
    \centering
    \begingroup
\normalsize
\setlength{\tabcolsep}{4mm}
\colorlet{mainresgroup}{black!10}
\arrayrulecolor{black}
\newcommand{\twoLineCell}[2]{\begin{tabular}[c]{@{}c@{}}#1\\#2\end{tabular}}
\begin{tabular}{l||cccccc||c}
\toprule
\textbf{Settings}
& \twoLineCell{\textbf{GQA}}{Acc. $\uparrow$}
& \twoLineCell{\textbf{MME}}{P+C $\uparrow$}
& \twoLineCell{\textbf{POPE}}{F1 $\uparrow$}
& \twoLineCell{\textbf{SQA}}{Acc. $\uparrow$}
& \twoLineCell{\textbf{VQA$^{\mathrm{t}}$}}{Acc. $\uparrow$}
& \twoLineCell{\textbf{OCR}}{Score $\uparrow$}
& \twoLineCell{\textbf{Avg.}}{$\uparrow$} \\
\midrule
\rowcolor{mainresgroup}
\multicolumn{1}{l}{\textbf{\textit{Full-token}}}
& \multicolumn{7}{c}{\textbf{\textit{576 visual tokens}}} \\
Vanilla & 61.90 & 1862 & 85.90 & 69.50 & 58.20 & 297 & 100.00\% \\
\midrule
\rowcolor{mainresgroup}
\multicolumn{1}{l}{\textbf{\textit{Prompt weight $\boldsymbol{\lambda}$}}}
& \multicolumn{7}{c}{\textbf{\textit{64 visual tokens}}} \\
$\lambda=0.0$ & 59.73 & 1747 & 84.31 & 68.22 & \textbf{56.56} & 282 & 96.46\% \\
$\lambda=0.5$ & \textbf{60.03} & 1738 & 84.61 & 68.82 & 56.31 & 277 & 96.31\% \\
$\lambda=1.0$ & 59.83 & 1748 & 84.96 & 69.06 & 56.55 & 282 & 96.82\% \\
$\lambda=1.5$ & 59.70 & 1761 & \textbf{85.70} & \textbf{69.20} & 56.40 & \textbf{285} & \textbf{97.20\%} \\
$\lambda=2.0$ & 59.32 & \textbf{1767} & 84.85 & 68.82 & 56.41 & 280 & 96.62\% \\
$\lambda=3.0$ & 59.35 & 1755 & 84.56 & 69.11 & 56.30 & 272 & 96.06\% \\
\midrule
\rowcolor{mainresgroup}
\multicolumn{1}{l}{\textbf{\textit{Feedback $\boldsymbol{\eta}$}}}
& \multicolumn{7}{c}{\textbf{\textit{64 visual tokens}}} \\
$\eta=0.2$ & 59.52 & \textbf{1775} & 84.74 & 68.86 & 56.46 & 277 & 96.58\% \\
$\eta=0.5$ & 59.45 & 1774 & 85.02 & 68.82 & 56.42 & 278 & 96.65\% \\
$\eta=0.8$ & \textbf{59.70} & 1761 & \textbf{85.70} & 69.20 & 56.40 & \textbf{285} & \textbf{97.20\%} \\
$\eta=1.0$ & 59.45 & 1760 & 84.70 & \textbf{69.51} & 56.50 & 279 & 96.70\% \\
$\eta=1.2$ & 59.59 & 1754 & 84.87 & 69.36 & \textbf{56.52} & 283 & 96.91\% \\
\midrule
\rowcolor{mainresgroup}
\multicolumn{1}{l}{\textbf{\textit{Top-$\boldsymbol{h}$ pooling}}}
& \multicolumn{7}{c}{\textbf{\textit{64 visual tokens}}} \\
$h=1$ & 59.55 & 1759 & 85.18 & 68.42 & \textbf{56.70} & 281 & 96.72\% \\
$h=3$ & \textbf{59.70} & 1761 & \textbf{85.70} & 69.20 & 56.40 & \textbf{285} & \textbf{97.20\%} \\
$h=5$ & 59.26 & \textbf{1782} & 84.92 & \textbf{69.21} & 56.50 & 280 & 96.87\% \\
$h=M$ & 59.61 & 1751 & 84.38 & 68.91 & 56.46 & 283 & 96.68\% \\
\bottomrule
\end{tabular}
\arrayrulecolor{black}
\endgroup

    \caption{Parameter sensitivity of DIVE on LLaVA-1.5-7B with 64 retained visual tokens. Each block varies one parameter while fixing the others at $\lambda=1.5$, $\eta=0.8$, and top-$h$ pooling ($h=3$). OCR denotes OCRBench, and Avg. is the mean of six unrounded benchmark-wise performance ratios relative to Vanilla.}
    \label{tab:parameter_sensitivity}
\end{table*}

Let $\ell$ denote the number of Transformer blocks completed before token selection; $\ell=2$ in all our experiments. For an $L$-layer LLM, let $F_{\leq\ell}(n)$ and $F_{>\ell}(n)$ denote the prefill costs of the layers up to the selection point and of the remaining layers, respectively, at prefix length $n$. With $T_{\mathrm{select}}=O\!\left(MNd+K((N+M)d+MN)\right)$, the LLM prefill costs are
\begin{equation}
    \begin{aligned}
        T_{\mathrm{Baseline}}
        &=F_{\leq\ell}(N+M)+F_{>\ell}(N+M),\\
        T_{\mathrm{DIVE}}
        &=F_{\leq\ell}(N+M)+T_{\mathrm{select}}
        +F_{>\ell}(K+M).
    \end{aligned}
\end{equation}
Under standard layer-wise KV caching, DIVE uses $O\!\left((\ell(N+M)+(L-\ell)(K+M))d\right)$ cache memory, compared with $O(L(N+M)d)$ for the baseline. If selection occurs before the first transformer layer ($\ell=0$), these expressions reduce to the simpler full-sequence formulas. The measured prefill latency reported in the main text includes the selection stage.

\section{Experimental Setup}
\label{app:exp_setup}
\subsection{Common evaluation protocol}
\noindent\textbf{Evaluation setup.}
All experiments were conducted in a zero-shot setting. For each benchmark, we followed its official evaluation split and protocol. Unless otherwise specified, evaluation used the default settings provided by \texttt{lmms-eval}. Within each comparison, all methods used the same evaluation data, evaluation scripts, and generation configuration.

\noindent\textbf{Comparison protocol.}
Within the same backbone and benchmark, token-reduction methods are compared at matched retained-token counts or retention ratios. The visual input is first constructed by the VLM pipeline, after which token selection is applied to the visual-token sequence. DIVE does not alter image resolution, video frame sampling, prompt tokens, or the language-model decoding procedure. After selection, the retained visual tokens are restored to their original order before being passed through the LLM layers. The vanilla baselines use the official implementations of the corresponding VLMs without token reduction.

\noindent\textbf{Benchmark metrics.}
For image benchmarks, we report accuracy for GQA, ScienceQA, TextVQA, and VizWiz; F1 score for POPE; the official Test-Dev score for VQAv2; the aggregate perception and cognition score for MME; and the official 1000-point score for OCRBench. For video benchmarks, we report multiple-choice accuracy on MVBench and VideoMME. For VideoMME, we report the overall result together with the short-, medium-, and long-duration splits.

For tables reporting relative average performance, let $s_{m,b}$ denote the score of method $m$ on benchmark metric $b$, and let $s_{\mathrm{full},b}$ denote the corresponding score of the full-token model with the same backbone. We compute
\begin{equation}
    \operatorname{Avg}(m)
    =
    \frac{1}{|\mathcal{B}|}
    \sum_{b \in \mathcal{B}}
    \frac{s_{m,b}}{s_{\mathrm{full},b}}
    \times 100\%,
    \label{eq:relative_average}
\end{equation}
where $\mathcal{B}$ is the set of benchmark metrics included in the corresponding table. All averages are computed from unrounded benchmark scores. Normalizing each metric before averaging prevents metrics with larger numerical scales, such as MME and OCRBench, from receiving disproportionate weight.

\subsection{Implementation details}
All experiments were conducted on a server with an Intel Xeon Gold 6230R CPU, 256 GB of system memory, and eight NVIDIA GeForce RTX 3090 GPUs (24 GB memory per GPU), running Ubuntu 20.04.6 LTS. Our implementation uses Python 3.10.20, PyTorch 2.1.2, and CUDA 12.1. All benchmark evaluations were conducted with \texttt{lmms-eval} v0.3.0 using a batch size of 1.

For all backbones, DIVE performs token selection after the first two LLM Transformer blocks, i.e., $\ell=2$. Here, $\ell$ denotes the number of completed blocks; equivalently, in our zero-based implementation, selection is applied before block 2 using the normalized output of block 1. Visual and valid prompt representations are gathered from this shared hidden-state matrix, with visual positions identified by the native multimodal input layout of each backbone. All textual and special tokens are preserved, and the retained visual tokens are restored to their original order before processing continues from block 2. This setting is fixed across backbones, model sizes, benchmarks, and token budgets. Competing methods follow their official implementations, including their original pruning locations or layer-wise schedules; comparisons are matched by backbone and retained-token budget rather than by pruning layer.

\subsection{Backbone architectures and token layouts}
We consider four representative open-weight LVLM families: LLaVA-1.5, LLaVA-NeXT, LLaVA-OV, and Qwen2-VL. They differ substantially in their visual-token layouts, including fixed-length image tokens, high-resolution multi-crop tokens, temporally ordered video tokens, and flexible-resolution variable-length tokens. These differences provide complementary settings for evaluating visual-token compression across diverse LVLM architectures.

\noindent\textbf{LLaVA-1.5.}
LLaVA-1.5~\cite{llava} is a fixed-resolution image LVLM built on the LLaVA architecture. Each input image is encoded into a fixed-length sequence of projected image-patch tokens, which provides a controlled setting for studying visual-token reduction under a stable token layout. The visual tokens occupy a clearly defined span in the multimodal input sequence, separate from text tokens and special delimiter tokens.

\noindent\textbf{LLaVA-NeXT.}
LLaVA-NeXT~\cite{llava-next} extends LLaVA-1.5 to high-resolution image inputs. Instead of representing an image with only a single fixed-resolution view, it combines a global image view with multiple local crops, producing a longer assembled visual-token sequence. This architecture is useful for evaluating whether token compression remains effective when visual evidence is distributed across both global context and fine-grained local regions.

\noindent\textbf{LLaVA-OV.}
LLaVA-OV~\cite{llava-ov} provides a video-capable extension of the LLaVA family. In the video setting used here, visual tokens are produced from multiple frames and arranged according to their temporal order. Compared with image-only backbones, this introduces a spatiotemporal visual sequence in which relevant evidence may appear across different frames as well as different spatial regions within each frame.

\noindent\textbf{Qwen2-VL.}
Qwen2-VL~\cite{qwen2-vl} supports flexible-resolution image and video inputs. Unlike fixed-token image backbones, its visual sequence length can vary across samples according to the input resolution and visual grid structure. The model provides grid metadata that identifies the visual-token layout, making it suitable for evaluating token compression under variable-length visual representations.

Together, these architectures cover fixed-resolution image tokens, high-resolution multi-crop image tokens, temporally ordered video tokens, and flexible-resolution variable-length visual tokens. This range allows the evaluation to test visual-token compression across substantially different LVLM token layouts rather than within a single backbone design.

\subsection{Experiment-specific settings}
Throughout these experiments, DIVE uses top-$h$ alignment pooling with $h=\min(3,M)$, prompt-conditioning strength $\lambda=1.5$, and feedback strength $\eta=0.8$.

\noindent\textbf{Image comparisons.}
LLaVA-1.5-7B and LLaVA-1.5-13B both take 576 visual tokens and retain $K\in\{192,128,64\}$ tokens. The 7B model is evaluated on GQA, MME, POPE, ScienceQA, TextVQA, OCRBench, VQAv2, and VizWiz; the 13B model uses the same benchmarks except VQAv2. LLaVA-NeXT-7B and LLaVA-NeXT-13B take 2880 high-resolution visual tokens and retain 640, 320, or 160 tokens. For Qwen2-VL-7B, we retain 33.3\% or 22.2\% of each variable-length visual sequence and evaluate ScienceQA, POPE, MME, and GQA.

\noindent\textbf{Video comparisons.}
We evaluate LLaVA-OV-7B and Qwen2-VL-7B on MVBench and VideoMME. LLaVA-OV-7B is tested at average retention ratios of 25\% and 15\%, while Qwen2-VL-7B is tested at 15\%. Frame sampling and input construction follow the default evaluator for each backbone.

\noindent\textbf{Efficiency evaluation.}
Table~\ref{tab:efficiency} evaluates fixed-budget efficiency on the complete POPE split with LLaVA-1.5-7B, comparing the 576-token vanilla model against 64-token compressed variants. Figure~\ref{fig:performance_latency_tradeoff} sweeps the retained-token budget on POPE, TextVQA, VizWiz, and GQA using the same backbone. Full-benchmark wall-clock time includes inference, prediction saving, and metric computation; prefill time is measured with CUDA events and includes token selection. All methods use one GPU and identical input order.

\noindent\textbf{Ablation and mechanism analysis.}
Table~\ref{tab:llava15_ablation} evaluates the main component ablation on LLaVA-1.5-7B with $K=64$ over GQA, MME, POPE, ScienceQA, TextVQA, and OCRBench. The \emph{w/o L-G}, \emph{w/o V-C}, and \emph{No Update} variants remove language guidance, visual-centric scoring, and iterative residual updates, respectively; all other settings remain fixed. Table~\ref{tab:ablation} repeats this comparison on LLaVA-NeXT-7B using 2880 high-resolution visual tokens and $K=320$.

The residual-feedback analysis uses LLaVA-1.5-7B on the complete MME split with $K=64$ and feedback schedules $\mathcal{R}=\{0,2,4,8,16,32,64\}$, with fixed initial scores, model, and sample order. Here, $r$ denotes the number of feedback updates, with intermediate updates distributed throughout the selection process. For sample $n$ and schedule $r$, let $S_{n,r}$ denote the retained set, let $\mathbf{x}_n$ denote the original visual-token representations, and let $\mathbf{q}_n$ denote the corresponding textual prompt. Let $\mathbf{y}_n=(y_{n,1},\ldots,y_{n,T_n})$ be the gold answer. The teacher-forced likelihood is
\begin{equation}
    L_{n,r}
    = \prod_{t=1}^{T_n}
    p_\theta(y_{n,t}\mid \mathbf{q}_n,y_{n,<t},\mathbf{x}_n,S_{n,r}).
    \label{eq:gold_answer_likelihood}
\end{equation}
We normalize this likelihood by its geometric mean across schedules:
\begin{equation}
    \begin{aligned}
        G_n
        &= \left(\prod_{r'\in\mathcal{R}}L_{n,r'}\right)^{1/|\mathcal{R}|}, \\
        \operatorname{RGL}_{n,r}
        &= \left(\frac{L_{n,r}}{G_n}-1\right)\times100\%.
    \end{aligned}
    \label{eq:relative_gold_likelihood}
\end{equation}

RGL is a post-selection diagnostic; higher values indicate stronger correct-answer support. Figure~\ref{fig:dynamic_feedback_analysis}(a) reports its distribution and the corresponding MME score.

For a retained set $S_{n,r}^{K}$ of size $K$, redundancy is the mean maximum cosine similarity among its original visual-token representations:
\begin{equation}
    \operatorname{Red}(S_{n,r}^{K})
    = \frac{1}{K}\sum_{i\in S_{n,r}^{K}}
    \max_{\substack{j\in S_{n,r}^{K}\\j\neq i}}
    \cos(\mathbf{x}_i,\mathbf{x}_j).
    \label{eq:retained_set_redundancy}
\end{equation}

Lower values indicate less within-set redundancy. The same schedules are compared at $K\in\{64,128,192\}$ in Figure~\ref{fig:dynamic_feedback_analysis}(b).

For runtime, each schedule retains 64 visual tokens and generates 32 tokens using greedy decoding, KV caching, and CUDA synchronization. Figure~\ref{fig:dynamic_feedback_analysis}(c) reports prefill overhead relative to No Update and decode-normalized latency, where each schedule's measured decoding time is replaced by the No Update mean to isolate the cost of feedback updates.

\subsection{Evaluation benchmarks}
We evaluate DIVE on 8 image benchmarks and 2 video benchmarks:

\begin{table*}[!t]
    \centering
    \begingroup
\setlength{\tabcolsep}{3.5mm}
\colorlet{singlegrp}{black!10}
\arrayrulecolor{black}
\begin{tabular}{@{}lcccccccc@{}}
\toprule
\textbf{Method} & \textbf{GQA} & \textbf{MME} & \textbf{POPE} & \textbf{SQA} & \textbf{VQA$^{\mathrm{t}}$} & \textbf{OCR} & \textbf{Average} & \textbf{Throughput} \\
\midrule
\rowcolor{singlegrp}
\multicolumn{9}{l}{\textbf{\textit{Full-token baseline (100\%)}}} \\
Vanilla & 64.3 & 1846 & 86.4 & 70.2 & 61.3 & 523 & 100.00\% & 1.00$\times$ \\
\midrule
\rowcolor{singlegrp}
\multicolumn{9}{c}{\textbf{\textit{Retain 320 tokens ($\downarrow$88.9\%)}}} \\
w/o L-G & \underline{59.3} & 1635 & 80.0 & \underline{68.6} & 54.6 & \underline{356} & \underline{88.04\%} & \underline{1.74$\times$} \\
w/o V-C & 58.8 & \underline{1642} & \underline{80.2} & \textbf{69.4} & \underline{55.2} & 329 & 87.51\% & 1.58$\times$ \\
No update & 52.4 & 1449 & 57.6 & 66.9 & 44.1 & 181 & 71.42\% & \textbf{1.94$\times$} \\
\textbf{DIVE} & \textbf{59.9} & \textbf{1769} & \textbf{83.5} & 68.2 & \textbf{56.0} & \textbf{377} & \textbf{91.04\%} & 1.56$\times$ \\
\bottomrule
\end{tabular}
\arrayrulecolor{black}
\endgroup

    \caption{Ablation study on LLaVA-NeXT-7B using the 2880-token high-resolution input and retaining 320 tokens. V-C and L-G denote the visual-centric and language-guided components, respectively. Avg. denotes the mean of the six metric-wise performance ratios relative to the vanilla model, and Thr. denotes throughput speedup.}
    \label{tab:ablation}
\end{table*}

\noindent\textbf{Image understanding benchmarks.}
\begin{itemize}

    \item \textbf{GQA}~\cite{GQA} evaluates compositional visual reasoning over real-world images, including questions about objects, attributes, and relations.

    \item \textbf{MME}~\cite{MME} evaluates perception- and cognition-oriented abilities, including visual recognition, OCR, commonsense understanding, and reasoning.

    \item \textbf{POPE}~\cite{POPE} evaluates object hallucination through binary questions about object existence in images.

    \item \textbf{ScienceQA}~\cite{ScienceQA} evaluates multimodal science question answering across natural, language, and social science topics.

    \item \textbf{TextVQA}~\cite{TextVQA} focuses on visual questions that require reading and reasoning over text in natural images.

    \item \textbf{OCRBench}~\cite{OCRBench} evaluates multimodal OCR ability across text recognition, scene-text VQA, document VQA, key information extraction, and handwritten mathematical expressions.

    \item \textbf{VQAv2}~\cite{VQAv2} evaluates open-ended visual question answering on diverse natural images using complementary image pairs.

    \item \textbf{VizWiz}~\cite{VizWiz} contains questions from blind users paired with user-captured images and human answers, often under noisy or difficult visual conditions.

\end{itemize}

\noindent\textbf{Video understanding benchmarks.}
\begin{itemize}

    \item \textbf{MVBench}~\cite{MVBench} evaluates video understanding through 20 multiple-choice tasks requiring temporal perception and reasoning.

    \item \textbf{VideoMME}~\cite{VideoMME} evaluates short-, medium-, and long-duration video understanding across diverse domains, with settings that may incorporate subtitles and audio.

\end{itemize}

\begin{table}[!t]
    \centering
    {\small
    \begingroup
\setlength{\tabcolsep}{1.6mm}
\colorlet{singlegrp}{black!10}
\colorlet{singleours}{orange!12}
\arrayrulecolor{black}
\begin{tabular}{@{}lccccc@{}}
\toprule
\textbf{Methods} & \textbf{SQA} & \textbf{POPE} & \textbf{MME} & \textbf{GQA} & \textbf{Avg.} \\
\midrule
\rowcolor{singlegrp}
\multicolumn{6}{l}{\textbf{\textit{Upper bound, all tokens (100\%)}}} \\
Qwen2-VL-7B & 85.5 & 87.8 & 2354 & 62.2 & 100.0\% \\
\midrule
\rowcolor{singlegrp}
\multicolumn{6}{c}{\textbf{\textit{Token reduction ($\downarrow$66.7\%)}}} \\
FastV [ECCV24] & \underline{82.2} & 84.8 & \underline{2280} & 58.8 & 96.0\% \\
DART [EMNLP25] & 80.2 & 85.2 & 2201 & 58.4 & 94.6\% \\
PDrop [CVPR25] & \underline{82.2} & 84.9 & \underline{2280} & 58.8 & 96.1\% \\
VisionZip [CVPR25] & \underline{82.2} & \underline{85.5} & 2275 & \underline{59.5} & \underline{96.5\%} \\
\textbf{DIVE [OURS]} & \textbf{82.4} & \textbf{87.9} & \textbf{2306} & \textbf{61.3} & \textbf{98.3\%} \\
\midrule
\rowcolor{singlegrp}
\multicolumn{6}{c}{\textbf{\textit{Token reduction ($\downarrow$77.8\%)}}} \\
FastV [ECCV24] & \underline{80.5} & 81.4 & 2208 & 56.4 & 92.8\% \\
DART [EMNLP25] & 78.6 & 82.8 & 2094 & 55.8 & 91.2\% \\
PDrop [CVPR25] & \underline{80.5} & 81.4 & 2208 & 56.4 & 92.8\% \\
VisionZip [CVPR25] & 79.9 & \underline{83.5} & \underline{2216} & \underline{57.3} & \underline{93.7\%} \\
\textbf{DIVE [OURS]} & \textbf{82.5} & \textbf{88.1} & \textbf{2268} & \textbf{60.7} & \textbf{97.7\%} \\
\bottomrule
\end{tabular}
\arrayrulecolor{black}
\endgroup
}
    \caption{Expanded comparison with Qwen2-VL-7B across multiple image-understanding benchmarks. Avg. denotes the mean of the unrounded per-benchmark performance ratios relative to the upper-bound model using all visual tokens.}
    \label{tab:qwen2vl_image_expanded}
\end{table}

\subsection{Comparison methods}
We compare DIVE with representative visual-token compression methods spanning mainstream pruning baselines and complementary greedy subset-selection methods, as follows:

\begin{table*}[!t]
    \centering
    {
    \begingroup
\normalsize
\setlength{\tabcolsep}{3.5mm}
\colorlet{mainresgroup}{black!10}
\colorlet{mainresours}{orange!12}
\arrayrulecolor{black}
\newcommand{\twoLineCell}[2]{\begin{tabular}[c]{@{}c@{}}#1\\#2\end{tabular}}
\begin{tabular}{lc||cccccc||c}
\toprule
\textbf{Method} & \textbf{Venue}
& \twoLineCell{\textbf{GQA}}{Acc. $\uparrow$}
& \twoLineCell{\textbf{MME}}{P+C $\uparrow$}
& \twoLineCell{\textbf{POPE}}{F1 $\uparrow$}
& \twoLineCell{\textbf{SQA}}{Acc. $\uparrow$}
& \twoLineCell{\textbf{VQA$^{\mathrm{t}}$}}{Acc. $\uparrow$}
& \twoLineCell{\textbf{OCR}}{Score $\uparrow$}
& \twoLineCell{\textbf{Avg.}}{$\uparrow$} \\
\midrule
\rowcolor{mainresgroup}
\multicolumn{2}{l}{\textbf{\textit{Full-token baseline}}} & \multicolumn{7}{c}{\textbf{\textit{576 visual tokens}}} \\
Vanilla & -- & 61.9 & 1862 & 85.9 & 69.5 & 58.2 & 297 & 100.00\% \\
\midrule
\rowcolor{mainresgroup}
\multicolumn{2}{l}{\textbf{\textit{Retain 192 visual tokens}}} & \multicolumn{7}{c}{\textbf{\textit{66.7\% token reduction}}} \\
MMTok & ICLR26 & 60.1 & 1774 & \underline{86.4} & 68.8 & \underline{57.7} & \underline{303} & 98.85\% \\
CDPruner & NeurIPS25 & \underline{60.5} & 1771 & \textbf{87.1} & 68.5 & 57.2 & \textbf{304} & 98.91\% \\
SCOPE & NeurIPS25 & 60.1 & \underline{1804} & 86.3 & \underline{68.8} & 57.6 & \underline{303} & \underline{99.07\%} \\
\textbf{DIVE} & -- & \textbf{61.4} & \textbf{1826} & \underline{86.4} & \textbf{69.1} & \textbf{58.0} & \textbf{304} & \textbf{99.88\%} \\
\midrule
\rowcolor{mainresgroup}
\multicolumn{2}{l}{\textbf{\textit{Retain 128 visual tokens}}} & \multicolumn{7}{c}{\textbf{\textit{77.8\% token reduction}}} \\
MMTok & ICLR26 & 59.3 & \underline{1780} & \underline{86.3} & \underline{68.8} & 57.0 & 300 & \underline{98.30\%} \\
CDPruner & NeurIPS25 & \underline{60.0} & 1743 & \textbf{87.1} & 68.4 & 56.4 & 298 & 97.93\% \\
SCOPE & NeurIPS25 & 59.6 & 1773 & 86.0 & 68.4 & \underline{57.2} & \underline{301} & 98.28\% \\
\textbf{DIVE} & -- & \textbf{60.8} & \textbf{1828} & 86.2 & \textbf{68.9} & \textbf{57.5} & \textbf{302} & \textbf{99.39\%} \\
\midrule
\rowcolor{mainresgroup}
\multicolumn{2}{l}{\textbf{\textit{Retain 64 visual tokens}}} & \multicolumn{7}{c}{\textbf{\textit{88.9\% token reduction}}} \\
MMTok & ICLR26 & 58.3 & \underline{1715} & \underline{85.8} & \textbf{69.2} & \underline{56.0} & \underline{283} & \underline{96.21\%} \\
CDPruner & NeurIPS25 & \underline{58.6} & 1710 & \textbf{87.3} & 68.2 & 55.2 & 271 & 95.39\% \\
SCOPE & NeurIPS25 & 58.3 & 1698 & 83.9 & \underline{68.6} & \textbf{56.4} & 282 & 95.60\% \\
\textbf{DIVE} & -- & \textbf{59.7} & \textbf{1761} & 85.7 & \textbf{69.2} & \textbf{56.4} & \textbf{285} & \textbf{97.20\%} \\
\bottomrule
\end{tabular}
\arrayrulecolor{black}
\endgroup
}
    \caption{Additional comparison with greedy coverage- and diversity-aware token-selection methods on LLaVA-1.5-7B under multiple visual-token budgets. All methods use the same backbone, retained-token budgets, evaluation protocol, and six benchmark metrics. Avg. is the mean of the six unrounded performance ratios relative to the full-token Vanilla baseline.}
    \label{tab:llava15_7b_recent_methods}
\end{table*}

\begin{itemize}

    \item \textbf{FastV}~\cite{FastV} performs visual token pruning by using attention information from early LLM layers to identify and discard less important visual tokens.

    \item \textbf{SparseVLM}~\cite{SparseVLM} sparsifies visual tokens according to cross-modal attention signals and introduces adaptive sparsity to reduce inference cost while retaining instruction-relevant evidence.

    \item \textbf{PDrop}~\cite{Pdrop} progressively drops visual tokens across model stages, forming a pyramid-like token structure that balances efficiency and performance.

    \item \textbf{VisionZip}~\cite{visionzip} compresses visual tokens by preserving compact informative tokens and reducing redundant visual context before language-model inference.

    \item \textbf{DART}~\cite{dart} adopts a duplication-aware pruning strategy, selecting visual tokens based on their redundancy relative to pivot tokens rather than relying only on standalone importance scores.

    \item \textbf{PruneSID}~\cite{PruneSID} groups visual tokens according to synergistic importance and diversity, then removes redundant tokens while preserving representative visual information. It further supports information-aware dynamic compression based on image complexity.

    \item \textbf{MMTok}~\cite{mmtok} formulates visual-token selection as multimodal maximum coverage. It greedily constructs a retained subset that jointly covers the textual tokens and the original visual-token set, thereby combining prompt relevance with visual representativeness.

    \item \textbf{CDPruner}~\cite{CDPruner} defines instruction-conditioned similarities between visual tokens and uses a determinantal point process to select a conditionally diverse subset. This formulation encourages the retained tokens to remain representative of the image while reducing instruction-irrelevant redundancy.

    \item \textbf{SCOPE}~\cite{SCOPE} combines visual saliency with the marginal coverage gain of each unselected token. It iteratively adds the candidate that contributes the largest saliency-weighted coverage gain and updates the coverage state of the retained set after each selection.

\end{itemize}

The main experiments compare DIVE with representative mainstream methods spanning attention-, relevance-, redundancy-, and importance--diversity-based pruning. We additionally evaluate MMTok, CDPruner, and SCOPE in Table~\ref{tab:llava15_7b_recent_methods} as complementary greedy or set-dependent subset-selection methods. For fair comparison, we follow the retained-token budgets or retention ratios used in the corresponding tables. All compared methods are evaluated on the same benchmark metrics within each table, and every reported average uses the same set of metrics.

\section{Additional Experiment Results}
\label{additional_experiment_results}

The main text summarizes DIVE's comparisons with representative mainstream visual-token pruning methods and its cross-backbone results. Here, we additionally compare with greedy subset-selection methods, provide the corresponding benchmark-level results, extend video evaluation to Qwen2-VL, and repeat the component and residual-update ablations. These experiments separate the effects of selection strategy, model family, input modality, model scale, and token budget.

\begin{table}[!t]
    \centering
    {\small
    \begingroup
\setlength{\tabcolsep}{0.65mm}
\colorlet{singlegrp}{black!10}
\colorlet{singleours}{orange!12}
\arrayrulecolor{black}
\begin{tabular}{@{}lcccccc@{}}
\toprule
\textbf{Methods} & \textbf{MVB} & \multicolumn{4}{c}{\textbf{VideoMME}} & \textbf{Avg.} \\
\cmidrule(lr){3-6}
 &  & Overall & Short & Medium & Long &  \\
\midrule
\rowcolor{singlegrp}
\multicolumn{7}{l}{\textbf{\textit{Upper bound, all tokens (100\%)}}} \\
Vanilla & 65.7 & 56.5 & 67.9 & 54.1 & 47.6 & 100.0\% \\
\midrule
\rowcolor{singlegrp}
\multicolumn{7}{c}{\textbf{\textit{Retention ratio = 15\%}}} \\
FastV [ECCV24] & 58.0 & 51.9 & 61.1 & 48.7 & \underline{45.8} & 91.3\% \\
DART [EMNLP25] & 57.6 & 52.2 & 60.8 & 50.3 & 45.6 & 91.7\% \\
\textbf{DIVE [OURS]} & \textbf{64.0} & \textbf{55.5} & \textbf{66.7} & \textbf{53.4} & \textbf{46.4} & \textbf{98.0\%} \\
\bottomrule
\end{tabular}
\arrayrulecolor{black}
\endgroup
}
    \caption{Performance comparison on Qwen2-VL-7B across video-understanding benchmarks.}
    \label{tab:qwen2vl_video}
\end{table}

\subsection{Additional Comparison with Greedy Subset Selection Methods}

The main experiments compare DIVE with representative mainstream visual-token pruning methods based on attention, prompt relevance, visual redundancy, and importance--diversity criteria. We further consider MMTok, CDPruner, and SCOPE, which construct retained sets through greedy coverage- or diversity-aware selection. Although these methods do not explicitly employ residual feedback, their subset construction makes later decisions depend on earlier selections and is therefore related in spirit to DIVE's \textit{select--update--re-evaluate} procedure. We include them as complementary comparisons to evaluate DIVE against a broader range of token-selection strategies.

DIVE achieves the highest relative average at all three budgets, retaining 99.88\%, 99.39\%, and 97.20\% of the full-token performance with 192, 128, and 64 retained tokens, respectively. It outperforms the strongest competing method by 0.81, 1.09, and 0.99 percentage points at the corresponding budgets. Across the individual benchmarks, DIVE leads or ties on five of the six metrics at each budget. These consistent results show that DIVE remains effective when evaluated alongside greedy coverage- and diversity-oriented token-selection methods.

\subsection{Additional Comparisons}
\noindent\textbf{Transfer to Qwen2-VL Image Inputs.}
The main text compares DIVE with FastV, DART, and VisionZip on Qwen2-VL; Table~\ref{tab:qwen2vl_image_expanded} additionally includes PDrop under the same budgets. DIVE leads all four benchmarks at both compression levels, retaining 98.3\% and 97.7\% of the all-token average at 66.7\% and 77.8\% token reduction. The margins over the strongest baseline increase from 1.8 to 4.0 percentage points as compression tightens. Together with Qwen2-VL's variable-length visual representation, this result supports transfer beyond scaling within LLaVA.

\noindent\textbf{Transfer to Qwen2-VL Video Inputs.}
Table~\ref{tab:qwen2vl_video} extends the video results beyond LLaVA-OV. With 15\% of the visual tokens retained, DIVE preserves 98.0\% of the all-token average, compared with 91.3\% for FastV and 91.7\% for DART. It leads every reported metric, with gains over the strongest baseline ranging from 0.6 points on long-duration VideoMME to 6.0 points on MVBench. The aggregate gain is therefore not confined to one temporal range.

\noindent\textbf{Scaling Across Model Sizes and Token Budgets.}
Tables~\ref{tab:dive_llava15_13b}--\ref{tab:dive_llavanext_13b} provide the per-benchmark results behind the main-text summary. On LLaVA-1.5-13B, DIVE retains 99.6\%, 99.1\%, and 97.4\% of the all-token average with 192, 128, and 64 tokens. Its margin over the strongest baseline grows from 1.6 to 4.2 percentage points as the budget tightens, consistent with selecting complementary evidence when each slot matters more. At 160 tokens on LLaVA-NeXT, DIVE retains 86.7\% on the 7B model and 87.2\% on the 13B model, leading the strongest baselines by 2.3 and 3.1 points. It leads five of seven task metrics on 7B and six of seven on 13B, so the advantage is not driven by a single benchmark.

\begin{table*}[!t]
    \centering
    {
    \begingroup
\normalsize
\setlength{\tabcolsep}{3mm}
\colorlet{mainresgroup}{black!10}
\arrayrulecolor{black}
\newcommand{\twoLineCell}[2]{\begin{tabular}[c]{@{}c@{}}#1\\#2\end{tabular}}
\newcommand{\methodCell}[2]{\begin{tabular}[c]{@{}l@{}}#1\\{\small #2}\end{tabular}}
\begin{tabular}{l||ccccccc||c}
\toprule
\textbf{Method}
& \twoLineCell{\textbf{GQA}}{Acc. $\uparrow$}
& \twoLineCell{\textbf{MME}}{P+C $\uparrow$}
& \twoLineCell{\textbf{POPE}}{F1 $\uparrow$}
& \twoLineCell{\textbf{SQA}}{Acc. $\uparrow$}
& \twoLineCell{\textbf{VQA$^{\mathrm{t}}$}}{Acc. $\uparrow$}
& \twoLineCell{\textbf{OCR}}{Score $\uparrow$}
& \twoLineCell{\textbf{VizWiz}}{Acc. $\uparrow$}
& \twoLineCell{\textbf{Avg.}}{$\uparrow$} \\
\midrule
\rowcolor{mainresgroup}
\multicolumn{9}{c}{\textbf{\textit{Upper bound, 576 visual tokens (100\%)}}} \\
\methodCell{LLaVA-1.5}{Vanilla-13B}
& \twoLineCell{63.3}{100.0\%} & \twoLineCell{1824}{100.0\%} & \twoLineCell{86.0}{100.0\%} & \twoLineCell{72.7}{100.0\%} & \twoLineCell{61.2}{100.0\%} & \twoLineCell{337}{100.0\%} & \twoLineCell{55.9}{100.0\%} & 100.0\% \\
\midrule
\rowcolor{mainresgroup}
\multicolumn{9}{c}{\textbf{\textit{Retain 192 visual tokens ($\downarrow$66.7\%)}}} \\
\methodCell{FastV}{\textit{ECCV24}}
& \twoLineCell{61.5}{97.2\%} & \twoLineCell{1798}{98.6\%} & \twoLineCell{82.9}{96.4\%} & \twoLineCell{73.8}{101.5\%} & \twoLineCell{60.4}{98.7\%} & \twoLineCell{320}{95.0\%} & \twoLineCell{55.0}{98.4\%} & 98.0\% \\
\methodCell{VisionZip}{\textit{CVPR25}}
& \twoLineCell{59.3}{93.7\%} & \twoLineCell{1809}{99.2\%} & \twoLineCell{84.8}{98.6\%} & \twoLineCell{73.7}{101.4\%} & \twoLineCell{59.6}{97.4\%} & \twoLineCell{336}{99.7\%} & \twoLineCell{52.4}{93.7\%} & 97.7\% \\
\methodCell{PruneSID}{\textit{ICLR26}}
& \twoLineCell{59.1}{93.4\%} & \twoLineCell{1782}{97.7\%} & \twoLineCell{85.4}{99.3\%} & \twoLineCell{72.9}{100.3\%} & \twoLineCell{42.4}{69.3\%} & \twoLineCell{307}{91.1\%} & \twoLineCell{53.8}{96.2\%} & 92.5\% \\
\methodCell{\textbf{DIVE}}{\textit{OURS}}
& \twoLineCell{62.6}{98.9\%} & \twoLineCell{1834}{100.5\%} & \twoLineCell{86.6}{100.7\%} & \twoLineCell{73.4}{101.0\%} & \twoLineCell{60.7}{99.2\%} & \twoLineCell{330}{97.9\%} & \twoLineCell{55.2}{98.7\%} & \textbf{99.6\%} \\
\midrule
\rowcolor{mainresgroup}
\multicolumn{9}{c}{\textbf{\textit{Retain 128 visual tokens ($\downarrow$77.8\%)}}} \\
\methodCell{FastV}{\textit{ECCV24}}
& \twoLineCell{59.6}{94.2\%} & \twoLineCell{1792}{98.2\%} & \twoLineCell{78.8}{91.6\%} & \twoLineCell{73.9}{101.7\%} & \twoLineCell{59.8}{97.7\%} & \twoLineCell{300}{89.0\%} & \twoLineCell{55.7}{99.6\%} & 96.0\% \\
\methodCell{VisionZip}{\textit{CVPR25}}
& \twoLineCell{57.6}{91.0\%} & \twoLineCell{1730}{94.8\%} & \twoLineCell{82.2}{95.6\%} & \twoLineCell{74.1}{101.9\%} & \twoLineCell{58.9}{96.2\%} & \twoLineCell{326}{96.7\%} & \twoLineCell{52.6}{94.1\%} & 95.8\% \\
\methodCell{PruneSID}{\textit{ICLR26}}
& \twoLineCell{58.2}{91.9\%} & \twoLineCell{1719}{94.2\%} & \twoLineCell{83.5}{97.1\%} & \twoLineCell{73.0}{100.4\%} & \twoLineCell{41.4}{67.6\%} & \twoLineCell{301}{89.3\%} & \twoLineCell{52.4}{93.7\%} & 90.6\% \\
\methodCell{\textbf{DIVE}}{\textit{OURS}}
& \twoLineCell{62.7}{99.1\%} & \twoLineCell{1833}{100.5\%} & \twoLineCell{86.8}{100.9\%} & \twoLineCell{72.6}{99.9\%} & \twoLineCell{60.3}{98.5\%} & \twoLineCell{322}{95.5\%} & \twoLineCell{55.5}{99.3\%} & \textbf{99.1\%} \\
\midrule
\rowcolor{mainresgroup}
\multicolumn{9}{c}{\textbf{\textit{Retain 64 visual tokens ($\downarrow$88.9\%)}}} \\
\methodCell{FastV}{\textit{ECCV24}}
& \twoLineCell{56.2}{88.8\%} & \twoLineCell{1684}{92.3\%} & \twoLineCell{70.5}{82.0\%} & \twoLineCell{72.7}{100.0\%} & \twoLineCell{56.6}{92.5\%} & \twoLineCell{276}{81.9\%} & \twoLineCell{55.5}{99.3\%} & 91.0\% \\
\methodCell{VisionZip}{\textit{CVPR25}}
& \twoLineCell{56.1}{88.6\%} & \twoLineCell{1687}{92.5\%} & \twoLineCell{75.9}{88.3\%} & \twoLineCell{74.3}{102.2\%} & \twoLineCell{57.5}{94.0\%} & \twoLineCell{309}{91.7\%} & \twoLineCell{53.4}{95.5\%} & 93.2\% \\
\methodCell{PruneSID}{\textit{ICLR26}}
& \twoLineCell{58.1}{91.8\%} & \twoLineCell{1750}{95.9\%} & \twoLineCell{82.9}{96.4\%} & \twoLineCell{72.2}{99.3\%} & \twoLineCell{41.2}{67.3\%} & \twoLineCell{287}{85.2\%} & \twoLineCell{54.1}{96.8\%} & 90.4\% \\
\methodCell{\textbf{DIVE}}{\textit{OURS}}
& \twoLineCell{61.8}{97.6\%} & \twoLineCell{1782}{97.7\%} & \twoLineCell{86.5}{100.6\%} & \twoLineCell{72.6}{99.9\%} & \twoLineCell{58.9}{96.2\%} & \twoLineCell{304}{90.2\%} & \twoLineCell{55.6}{99.5\%} & \textbf{97.4\%} \\
\bottomrule
\end{tabular}
\arrayrulecolor{black}
\endgroup
}
    \caption{Additional comparison on LLaVA-1.5-13B with DIVE under multiple visual-token budgets. The first and second lines of each metric cell report the raw score and percentage relative to the all-token Vanilla upper bound, respectively. Avg. is the mean of the seven unrounded relative percentages.}
    \label{tab:dive_llava15_13b}
\end{table*}

\begin{figure*}[!t]
    \centering
    \includegraphics[width=0.7\textwidth]{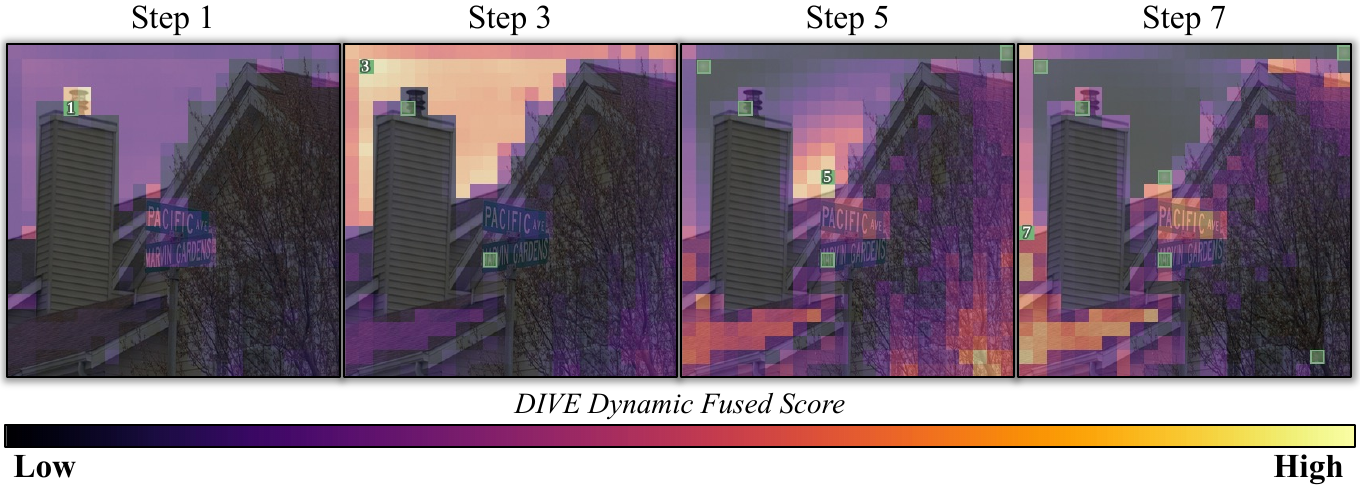}
    \caption{\textbf{Evolution of DIVE's dynamic fused score.}
    Numbered green boxes mark tokens selected by the displayed steps. The score
    map is recomputed as the retained set grows, redirecting later selections
    toward visual content not yet represented.}
    \label{fig:dynamic_score_evolution}
\end{figure*}

\begin{table*}[!t]
    \centering
    {
    \begingroup
\normalsize
\setlength{\tabcolsep}{3mm}
\colorlet{mainresgroup}{black!10}
\arrayrulecolor{black}
\newcommand{\twoLineCell}[2]{\begin{tabular}[c]{@{}c@{}}#1\\#2\end{tabular}}
\newcommand{\methodCell}[2]{\begin{tabular}[c]{@{}l@{}}#1\\{\small #2}\end{tabular}}
\begin{tabular}{l||ccccccc||c}
\toprule
\textbf{Method}
& \twoLineCell{\textbf{GQA}}{Acc. $\uparrow$}
& \twoLineCell{\textbf{MME}}{P+C $\uparrow$}
& \twoLineCell{\textbf{POPE}}{F1 $\uparrow$}
& \twoLineCell{\textbf{SQA}}{Acc. $\uparrow$}
& \twoLineCell{\textbf{VQA$^{\mathrm{t}}$}}{Acc. $\uparrow$}
& \twoLineCell{\textbf{OCR}}{Score $\uparrow$}
& \twoLineCell{\textbf{VizWiz}}{Acc. $\uparrow$}
& \twoLineCell{\textbf{Avg.}}{$\uparrow$} \\
\midrule
\rowcolor{mainresgroup}
\multicolumn{9}{c}{\textbf{\textit{Upper bound, 2880 visual tokens (100\%)}}} \\
\methodCell{LLaVA-NeXT}{Vanilla-7B}
& \twoLineCell{64.3}{100.0\%} & \twoLineCell{1846}{100.0\%} & \twoLineCell{86.4}{100.0\%} & \twoLineCell{70.2}{100.0\%} & \twoLineCell{61.3}{100.0\%} & \twoLineCell{523}{100.0\%} & \twoLineCell{58.6}{100.0\%} & 100.0\% \\
\midrule
\rowcolor{mainresgroup}
\multicolumn{9}{c}{\textbf{\textit{Retain 640 visual tokens ($\downarrow$77.8\%)}}} \\
\methodCell{FastV}{\textit{ECCV24}}
& \twoLineCell{59.2}{92.1\%} & \twoLineCell{1712}{92.7\%} & \twoLineCell{75.8}{87.7\%} & \twoLineCell{68.3}{97.3\%} & \twoLineCell{56.2}{91.7\%} & \twoLineCell{395}{75.5\%} & \twoLineCell{51.9}{88.6\%} & 89.4\% \\
\methodCell{VisionZip}{\textit{CVPR25}}
& \twoLineCell{59.8}{93.0\%} & \twoLineCell{1780}{96.4\%} & \twoLineCell{77.4}{89.6\%} & \twoLineCell{68.3}{97.3\%} & \twoLineCell{51.1}{83.4\%} & \twoLineCell{319}{61.0\%} & \twoLineCell{54.5}{93.0\%} & 87.7\% \\
\methodCell{PruneSID}{\textit{ICLR26}}
& \twoLineCell{62.2}{96.7\%} & \twoLineCell{1743}{94.4\%} & \twoLineCell{84.1}{97.3\%} & \twoLineCell{66.6}{94.9\%} & \twoLineCell{56.5}{92.2\%} & \twoLineCell{420}{80.3\%} & \twoLineCell{53.9}{92.0\%} & 92.5\% \\
\methodCell{\textbf{DIVE}}{\textit{OURS}}
& \twoLineCell{62.3}{96.9\%} & \twoLineCell{1791}{97.0\%} & \twoLineCell{83.6}{96.8\%} & \twoLineCell{68.4}{97.4\%} & \twoLineCell{56.9}{92.8\%} & \twoLineCell{447}{85.5\%} & \twoLineCell{54.3}{92.7\%} & \textbf{94.2\%} \\
\midrule
\rowcolor{mainresgroup}
\multicolumn{9}{c}{\textbf{\textit{Retain 320 visual tokens ($\downarrow$88.9\%)}}} \\
\methodCell{FastV}{\textit{ECCV24}}
& \twoLineCell{55.7}{86.6\%} & \twoLineCell{1606}{87.0\%} & \twoLineCell{68.8}{79.6\%} & \twoLineCell{69.1}{98.4\%} & \twoLineCell{54.4}{88.7\%} & \twoLineCell{334}{63.9\%} & \twoLineCell{51.2}{87.4\%} & 84.5\% \\
\methodCell{VisionZip}{\textit{CVPR25}}
& \twoLineCell{59.3}{92.2\%} & \twoLineCell{1708}{92.5\%} & \twoLineCell{82.4}{95.4\%} & \twoLineCell{67.9}{96.7\%} & \twoLineCell{55.1}{89.9\%} & \twoLineCell{369}{70.6\%} & \twoLineCell{52.6}{89.8\%} & 89.6\% \\
\methodCell{PruneSID}{\textit{ICLR26}}
& \twoLineCell{60.0}{93.3\%} & \twoLineCell{1743}{94.4\%} & \twoLineCell{84.6}{97.9\%} & \twoLineCell{68.1}{97.0\%} & \twoLineCell{53.3}{86.9\%} & \twoLineCell{362}{69.2\%} & \twoLineCell{52.2}{89.1\%} & 89.7\% \\
\methodCell{\textbf{DIVE}}{\textit{OURS}}
& \twoLineCell{59.9}{93.2\%} & \twoLineCell{1769}{95.8\%} & \twoLineCell{83.5}{96.6\%} & \twoLineCell{68.2}{97.2\%} & \twoLineCell{56.0}{91.4\%} & \twoLineCell{377}{72.1\%} & \twoLineCell{52.9}{90.3\%} & \textbf{90.9\%} \\
\midrule
\rowcolor{mainresgroup}
\multicolumn{9}{c}{\textbf{\textit{Retain 160 visual tokens ($\downarrow$94.4\%)}}} \\
\methodCell{FastV}{\textit{ECCV24}}
& \twoLineCell{50.3}{78.2\%} & \twoLineCell{1453}{78.7\%} & \twoLineCell{47.9}{55.4\%} & \twoLineCell{68.5}{97.6\%} & \twoLineCell{50.1}{81.7\%} & \twoLineCell{261}{49.9\%} & \twoLineCell{49.2}{84.0\%} & 75.1\% \\
\methodCell{VisionZip}{\textit{CVPR25}}
& \twoLineCell{55.6}{86.5\%} & \twoLineCell{1676}{90.8\%} & \twoLineCell{72.5}{83.9\%} & \twoLineCell{68.0}{96.9\%} & \twoLineCell{50.3}{82.1\%} & \twoLineCell{300}{57.4\%} & \twoLineCell{54.6}{93.2\%} & 84.4\% \\
\methodCell{PruneSID}{\textit{ICLR26}}
& \twoLineCell{56.9}{88.5\%} & \twoLineCell{1563}{84.7\%} & \twoLineCell{79.8}{92.4\%} & \twoLineCell{68.0}{96.9\%} & \twoLineCell{45.5}{74.2\%} & \twoLineCell{300}{57.4\%} & \twoLineCell{50.0}{85.3\%} & 82.8\% \\
\methodCell{\textbf{DIVE}}{\textit{OURS}}
& \twoLineCell{58.7}{91.3\%} & \twoLineCell{1658}{89.8\%} & \twoLineCell{80.2}{92.8\%} & \twoLineCell{69.2}{98.6\%} & \twoLineCell{53.2}{86.8\%} & \twoLineCell{310}{59.3\%} & \twoLineCell{51.9}{88.6\%} & \textbf{86.7\%} \\
\bottomrule
\end{tabular}
\arrayrulecolor{black}
\endgroup
}
    \caption{Additional comparison on LLaVA-NeXT-7B with DIVE under multiple visual-token budgets. The first and second lines of each metric cell report the raw score and percentage relative to the all-token Vanilla upper bound, respectively. Avg. is the mean of the seven unrounded relative percentages.}
    \label{tab:dive_llavanext_7b}
\end{table*}

\begin{table*}[!t]
    \centering
    {
    \begingroup
\normalsize
\setlength{\tabcolsep}{3mm}
\colorlet{mainresgroup}{black!10}
\arrayrulecolor{black}
\newcommand{\twoLineCell}[2]{\begin{tabular}[c]{@{}c@{}}#1\\#2\end{tabular}}
\newcommand{\methodCell}[2]{\begin{tabular}[c]{@{}l@{}}#1\\{\small #2}\end{tabular}}
\begin{tabular}{l||ccccccc||c}
\toprule
\textbf{Method}
& \twoLineCell{\textbf{GQA}}{Acc. $\uparrow$}
& \twoLineCell{\textbf{MME}}{P+C $\uparrow$}
& \twoLineCell{\textbf{POPE}}{F1 $\uparrow$}
& \twoLineCell{\textbf{SQA}}{Acc. $\uparrow$}
& \twoLineCell{\textbf{VQA$^{\mathrm{t}}$}}{Acc. $\uparrow$}
& \twoLineCell{\textbf{OCR}}{Score $\uparrow$}
& \twoLineCell{\textbf{VizWiz}}{Acc. $\uparrow$}
& \twoLineCell{\textbf{Avg.}}{$\uparrow$} \\
\midrule
\rowcolor{mainresgroup}
\multicolumn{9}{c}{\textbf{\textit{Upper bound, 2880 visual tokens (100\%)}}} \\
\methodCell{LLaVA-NeXT}{Vanilla-13B}
& \twoLineCell{65.4}{100.0\%} & \twoLineCell{1892}{100.0\%} & \twoLineCell{86.3}{100.0\%} & \twoLineCell{73.6}{100.0\%} & \twoLineCell{64.2}{100.0\%} & \twoLineCell{551}{100.0\%} & \twoLineCell{61.9}{100.0\%} & 100.0\% \\
\midrule
\rowcolor{mainresgroup}
\multicolumn{9}{c}{\textbf{\textit{Retain 640 visual tokens ($\downarrow$77.8\%)}}} \\
\methodCell{FastV}{\textit{ECCV24}}
& \twoLineCell{62.0}{94.8\%} & \twoLineCell{1847}{97.6\%} & \twoLineCell{78.3}{90.7\%} & \twoLineCell{72.7}{98.8\%} & \twoLineCell{60.4}{94.1\%} & \twoLineCell{405}{73.5\%} & \twoLineCell{56.0}{90.5\%} & 91.4\% \\
\methodCell{VisionZip}{\textit{CVPR25}}
& \twoLineCell{61.8}{94.5\%} & \twoLineCell{1816}{96.0\%} & \twoLineCell{80.0}{92.7\%} & \twoLineCell{71.4}{97.0\%} & \twoLineCell{54.2}{84.4\%} & \twoLineCell{326}{59.2\%} & \twoLineCell{54.3}{87.7\%} & 87.4\% \\
\methodCell{PruneSID}{\textit{ICLR26}}
& \twoLineCell{63.1}{96.5\%} & \twoLineCell{1837}{97.1\%} & \twoLineCell{84.5}{97.9\%} & \twoLineCell{72.9}{99.0\%} & \twoLineCell{60.2}{93.8\%} & \twoLineCell{449}{81.5\%} & \twoLineCell{53.5}{86.4\%} & 93.2\% \\
\methodCell{\textbf{DIVE}}{\textit{OURS}}
& \twoLineCell{62.7}{95.9\%} & \twoLineCell{1860}{98.3\%} & \twoLineCell{83.7}{97.0\%} & \twoLineCell{72.3}{98.2\%} & \twoLineCell{61.3}{95.5\%} & \twoLineCell{463}{82.2\%} & \twoLineCell{58.8}{94.0\%} & \textbf{94.4\%} \\
\midrule
\rowcolor{mainresgroup}
\multicolumn{9}{c}{\textbf{\textit{Retain 320 visual tokens ($\downarrow$88.9\%)}}} \\
\methodCell{FastV}{\textit{ECCV24}}
& \twoLineCell{58.5}{89.4\%} & \twoLineCell{1810}{95.7\%} & \twoLineCell{72.3}{83.8\%} & \twoLineCell{72.1}{98.0\%} & \twoLineCell{57.8}{90.0\%} & \twoLineCell{344}{62.4\%} & \twoLineCell{54.0}{87.2\%} & 86.7\% \\
\methodCell{VisionZip}{\textit{CVPR25}}
& \twoLineCell{60.7}{92.8\%} & \twoLineCell{1804}{95.3\%} & \twoLineCell{78.5}{91.0\%} & \twoLineCell{71.9}{97.7\%} & \twoLineCell{52.7}{82.1\%} & \twoLineCell{320}{58.1\%} & \twoLineCell{54.8}{88.5\%} & 86.5\% \\
\methodCell{PruneSID}{\textit{ICLR26}}
& \twoLineCell{61.0}{93.3\%} & \twoLineCell{1781}{94.1\%} & \twoLineCell{84.4}{97.8\%} & \twoLineCell{72.2}{98.1\%} & \twoLineCell{56.1}{87.4\%} & \twoLineCell{379}{68.8\%} & \twoLineCell{51.8}{83.7\%} & 89.0\% \\
\methodCell{\textbf{DIVE}}{\textit{OURS}}
& \twoLineCell{60.7}{92.8\%} & \twoLineCell{1837}{97.1\%} & \twoLineCell{83.1}{96.3\%} & \twoLineCell{72.7}{98.8\%} & \twoLineCell{58.7}{91.4\%} & \twoLineCell{426}{71.9\%} & \twoLineCell{55.9}{90.3\%} & \textbf{91.2\%} \\
\midrule
\rowcolor{mainresgroup}
\multicolumn{9}{c}{\textbf{\textit{Retain 160 visual tokens ($\downarrow$94.4\%)}}} \\
\methodCell{FastV}{\textit{ECCV24}}
& \twoLineCell{53.5}{81.8\%} & \twoLineCell{1669}{88.2\%} & \twoLineCell{60.8}{70.5\%} & \twoLineCell{72.0}{97.8\%} & \twoLineCell{53.8}{83.8\%} & \twoLineCell{251}{45.6\%} & \twoLineCell{50.4}{81.4\%} & 78.4\% \\
\methodCell{VisionZip}{\textit{CVPR25}}
& \twoLineCell{58.5}{89.4\%} & \twoLineCell{1681}{88.8\%} & \twoLineCell{75.2}{87.1\%} & \twoLineCell{71.0}{96.5\%} & \twoLineCell{54.0}{84.1\%} & \twoLineCell{309}{56.1\%} & \twoLineCell{53.5}{86.4\%} & 84.1\% \\
\methodCell{PruneSID}{\textit{ICLR26}}
& \twoLineCell{56.6}{86.5\%} & \twoLineCell{1578}{83.4\%} & \twoLineCell{73.8}{85.5\%} & \twoLineCell{71.8}{97.6\%} & \twoLineCell{44.3}{69.0\%} & \twoLineCell{283}{51.4\%} & \twoLineCell{47.5}{76.7\%} & 78.6\% \\
\methodCell{\textbf{DIVE}}{\textit{OURS}}
& \twoLineCell{59.3}{90.7\%} & \twoLineCell{1760}{93.0\%} & \twoLineCell{81.2}{94.1\%} & \twoLineCell{72.4}{98.4\%} & \twoLineCell{56.7}{88.3\%} & \twoLineCell{359}{59.7\%} & \twoLineCell{53.4}{86.3\%} & \textbf{87.2\%} \\
\bottomrule
\end{tabular}
\arrayrulecolor{black}
\endgroup
}
    \caption{Additional comparison on LLaVA-NeXT-13B with DIVE under multiple visual-token budgets. The first and second lines of each metric cell report the raw score and percentage relative to the all-token Vanilla upper bound, respectively. Avg. is the mean of the seven unrounded relative percentages.}
    \label{tab:dive_llavanext_13b}
\end{table*}

\noindent\textbf{Parameter Sensitivity.}
Across all values of $\lambda$, $\eta$, and $h$ evaluated in Table~\ref{tab:parameter_sensitivity}, Avg. remains between 96.06\% and 97.20\%, showing that DIVE is stable over a broad range of parameter settings.

\subsection{Ablation and Feedback Analysis}

\noindent\textbf{Component and Residual-Update Ablations.}
The main text isolates the visual-centric (V-C) and language-guided (L-G) signals on LLaVA-1.5; Table~\ref{tab:ablation} repeats the comparison on LLaVA-NeXT-7B. Under the 2880-to-320-token setting, DIVE retains 91.04\% of the full-token average across the six reported benchmarks. Removing L-G and V-C lowers the relative average to 88.04\% and 87.51\%, respectively. The No Update variant is faster but drops to 71.42\%, showing that the initial ranking alone is not sufficient when high-resolution visual evidence is compressed aggressively. This controlled comparison supports residual re-evaluation while making its throughput cost explicit.

\noindent\textbf{Cost of Feedback Updates.}
Main-text Figure~\ref{fig:dynamic_feedback_analysis}(c) reports the update-frequency cost on the complete MME split with a 64-token budget and 32 generated tokens. Relative to No Update, Full Update adds 13.6\,ms (30.7\%) to mean prefill latency. To remove decode-side runtime variation unrelated to feedback, we normalize every schedule to the No Update mean decode latency; the resulting end-to-end latency changes from 762.1 to 775.8\,ms (1.8\%). The additional cost is concentrated in the prefill stage under this protocol.

\section{Additional Visualization Results}
\label{sec:additional_visualization_results}

Figure~\ref{fig:dynamic_score_evolution} visualizes the fused score during
iterative selection. Once a token is selected, high-score regions change in
subsequent steps as the residual state discounts evidence already represented
by the retained set. The example illustrates the set-dependent behavior
quantified by the main-text likelihood and redundancy analyses.

\begin{figure*}[!t]
    \centering
    \includegraphics[width=0.9\textwidth]{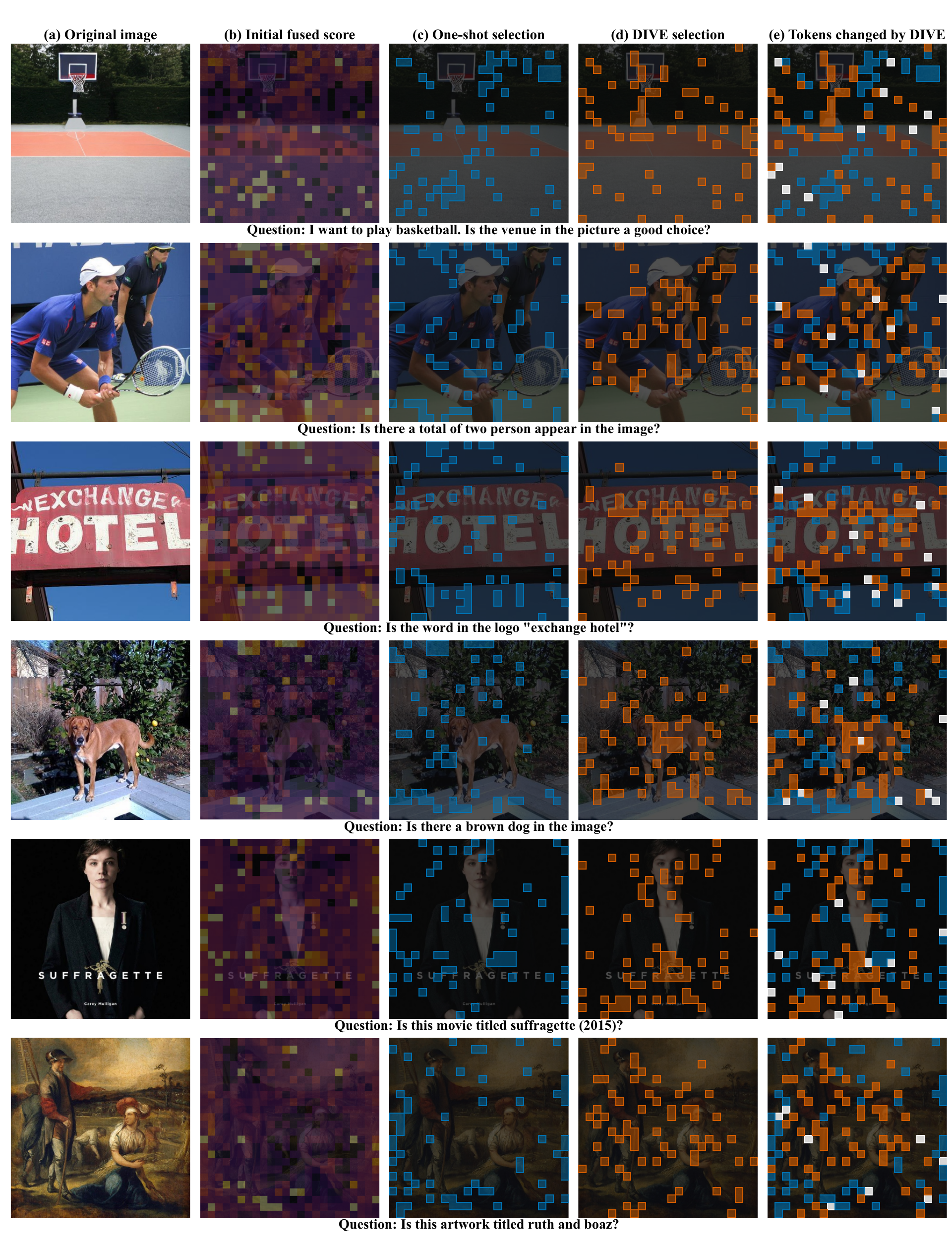}
    \caption{Qualitative comparison of prompt-conditioned visual-token selection
    across original images, initial fused scores, one-shot selection, DIVE
    selection, and their token differences. Blue, orange, and white marks denote
    one-shot-only, DIVE-only, and shared tokens, respectively; DIVE shifts
    selections toward complementary, query-relevant regions through residual
    re-evaluation.}
    \label{fig:dive_token_selection_gallery}
\end{figure*}

Figure~\ref{fig:dive_token_selection_gallery} provides a qualitative view of
how the prompt-conditioned token selectors distribute their retained tokens.
The examples cover different visual layouts and question types, making the
contrast between fixed one-shot ranking and DIVE's residual-aware selection
directly visible.

\begin{figure*}[!t]
    \centering
    \includegraphics[width=0.9\textwidth]{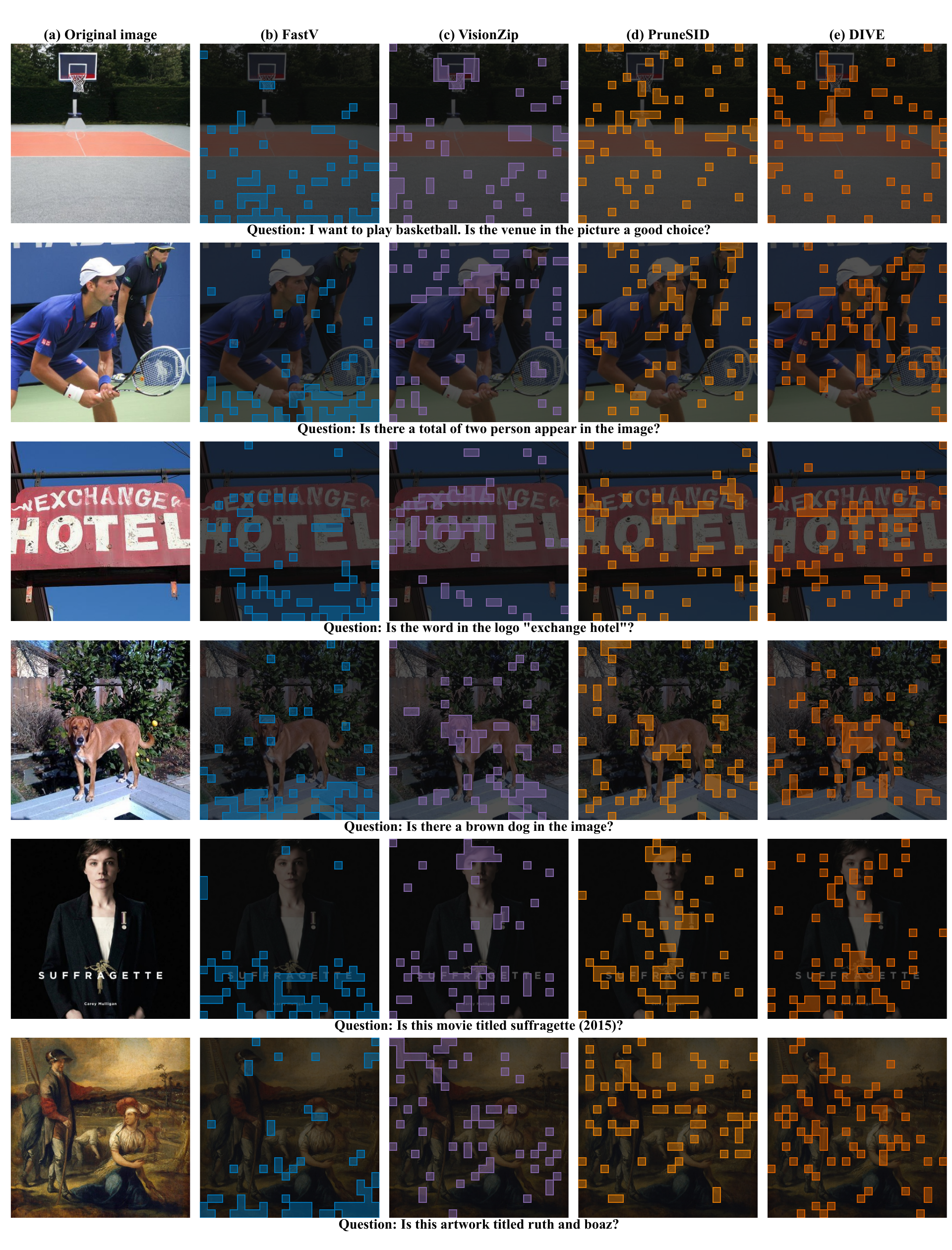}
    \caption{Qualitative comparison of real visual-token masks produced by
    FastV, VisionZip, PruneSID, and DIVE. Colored squares mark each
    method's retained visual tokens.}
    \label{fig:method_token_selection_comparison}
\end{figure*}

We further compare the spatial distributions produced by representative
token-pruning strategies in Figure~\ref{fig:method_token_selection_comparison}.
The examples are chosen to highlight cases in which DIVE recovers the correct
answer while the static selector does not.

\section{Limitations}
Two limitations warrant further investigation. First, DIVE operates on intermediate visual-token representations and therefore assumes access to model internals. This requirement currently restricts its application to open-weight MLLMs; extending DIVE to API-only systems would require selection signals that can be obtained without direct access to hidden states or inference procedures. Second, the amount of removable visual redundancy may depend on the underlying architecture. Models whose visual encoders already produce relatively compact representations may leave less redundancy for an additional pruning stage and may therefore require less aggressive compression. Future work will explore architecture-adaptive token budgets and black-box-compatible efficiency methods for heterogeneous visual encoding pipelines.

\end{document}